\documentclass{article}
\newcommand*{\RootPath}{./}%

\usepackage[utf8]{inputenc} %
\usepackage[T1]{fontenc}    %
\usepackage{nicefrac}       %
\usepackage{hyperref}
\usepackage{url}
\usepackage{booktabs}       %
\usepackage{amsthm}
\usepackage{amsfonts}       %
\usepackage{nicefrac}       %
\usepackage{microtype}      %
\usepackage{xcolor}         %
\usepackage{multirow}
\usepackage[
backend=biber,
natbib=true,
style=numeric,
maxcitenames=1
]{biblatex}
\usepackage{graphicx}
\graphicspath{ {\RootPath/} }
\usepackage{subcaption}
\usepackage{empheq}
\usepackage{colortbl}
\usepackage{color}
\definecolor{gray}{rgb}{0.2,0.2,0.3}

\usepackage{amsmath}
\usepackage{makecell}

\usepackage{dsfont}
\usepackage{tikz}
\usepackage{xspace}
\usepackage{tabularx}
\newcommand{\ie}{\textit{i.e.}\xspace}
\newcommand{\eg}{\textit{e.g.}\xspace}
\usepackage{algorithm}
\usepackage[noend]{algpseudocode}
\usepackage{bbm}
\usepackage{numprint}
\usepackage{cleveref}
\usepackage{wrapfig}
\usepackage{pifont}
\usepackage{xspace}
\usepackage{tabularx}
\usepackage{makecell}
\usepackage{enotez}
\usepackage{setspace}
\usepackage{multicol}
\usepackage{comment}
\usepackage{enumitem}
\usepackage{fancyhdr}
\usepackage{amsthm}

\usepackage{authblk}
\usepackage{fullpage}
\usepackage{amsthm}

\newtheorem{theorem}{Theorem}

\newif\ifdraft

\draftfalse

\newcommand{\dtodo}[1]{}

\ifdraft
    \usepackage[textsize=tiny, backgroundcolor=white]{todonotes}
    \newcommand{\my}[1]{\textcolor{teal}{M: #1}}
    \newcommand{\franzi}[1]{\textcolor{violet}{franzi: #1}}
    \newcommand{\ilia}[1]{\textcolor{green}{ilia: #1}}
    \newcommand{\sierra}[1]{\textcolor{blue}{sierra: #1}}
    \newcommand{\jonas}[1]{\textcolor{olive}{jonas:#1}}
    \newcommand{\patty}[1]{\textcolor{cyan}{patty:#1}}
    
    \newcommand{\varun}[1]{\todo{\textcolor{red}{VC:#1}}}
    \newcommand{\nicolas}[1]{\textcolor{olive}{NP:#1}}
    \newcommand{\stephan}[1]{\todo{\textcolor{red}{SR: #1}}}
    
    \newcommand{\anvith}[1]{\textcolor{brown}{2D:#1}}
    \newcommand{\ahmad}[1]{\textcolor{lime}{ahmad:#1}}
    \newcommand{\nick}[1]{\textcolor{yellow}{HJ:#1}}
    \newcommand{\tf}[1]{\todo{{\color{black}F: #1}}}

    \DeclareCaptionFont{blue}{\color{blue}}
    \DeclareCaptionFont{black}{\color{black}}

    \newenvironment{newlyaddedEnv}{
    \color{blue}
    \everymath{\color{blue}}
    \everydisplay{\color{blue}}
    \captionsetup{font=blue}
    }{
    \color{black}
    \everymath{\color{black}}
    \everydisplay{\color{black}}
    \captionsetup{font=black}
    }

\else
    \newcommand{\todo}[1]{}
    \newcommand{\my}[1]{}
    \newcommand{\franzi}[1]{}
    \newcommand{\ilia}[1]{}
    \newcommand{\anvith}[1]{}
    \newcommand{\sierra}[1]{}
    \newcommand{\jonas}[1]{}
    \newcommand{\patty}[1]{}
    
    \newcommand{\varun}[1]{}
    \newcommand{\nicolas}[1]{}
    \newcommand{\stephan}[1]{}
    \newcommand{\daivd}[1]{}
    \newcommand{\ahmad}[1]{}
    \newcommand{\nick}[1]{}
    \newcommand{\tf}[1]{}

\fi

\makeatletter
\newcommand{\rpriv}[1][\@nil]{%
    \def\tmp{#1}%
    \ifx\tmp\@nnil
        \varepsilon
    \else
        #1
    \fi}

\newcommand{\rfair}[1][\@nil]{%
    \def\tmp{#1}%
    \ifx\tmp\@nnil
        \gamma
    \else
        #1
    \fi}

\newcommand{\tpriv}[1][\@nil]{%
    \def\tmp{#1}%
    \ifx\tmp\@nnil
        \tau_\priv
    \else
        #1
    \fi}

\newcommand{\tfair}[1][\@nil]{%
    \def\tmp{#1}%
    \ifx\tmp\@nnil
        \tau_\build
    \else
        #1
    \fi}

\makeatother

\newcommand{\fair}{\textsl{fair}}
\newcommand{\priv}{\textsl{priv}}
\newcommand{\build}{\textsl{build}}
\newcommand{\reg}{\textsl{reg}}
\newcommand{\model}{\omega}

\newcommand{\cpriv}{C_\priv}
\newcommand{\cfair}{C_\fair}

\newcommand{\buildPriv}{\lambda_\priv}
\newcommand{\buildFair}{\lambda_\fair}

\newcommand{\eps}{\varepsilon}

\usepackage{cancel}
\usepackage{dsfont}
\usepackage{tikz}
\usepackage{xspace}
\usepackage{tabularx}
\usepackage{algorithm}
\usepackage[noend]{algpseudocode}

\usepackage{numprint}
\usepackage{cleveref}

\DeclareMathOperator*{\argmin}{arg\,min}

\algnewcommand\algorithmicinput{\textbf{Input:}}
\algnewcommand\Input{\item[\algorithmicinput]}%

\newcommand{\Deval}{D_\text{eval}}

\newcommand{\testdataset}{D_\text{test}}

\newcommand{\posReals}{\mathbb{R}^+}

\newcommand{\gmm}{\gamma}

\newcommand{\heps}{\widehat{\eps}}
\newcommand{\hgmm}{\widehat{\gmm}}

\newcommand{\acc}{\text{acc}}
\newcommand{\err}{\text{err}}
\newcommand{\cov}{\text{cov}}

\newcommand{\pf}{Pareto frontier\xspace}

\newcommand{\pfs}{Pareto frontiers\xspace}
\newcommand{\pe}{Pareto-efficient\xspace}

\newcommand{\fpate}{FairPATE\xspace}
\newcommand{\fDPbl}{DPSGD-Global-Adapt}

\newcommand{\pp}{\textsc{ParetoPlay}\xspace}
\newcommand{\specg}{\textsc{SpecGame}\xspace}

\newcommand{\bs}{\boldsymbol{s}} %

\newcommand{\cS}{\mathcal{S}} %
\newcommand{\cA}{\mathcal{A}} %
\newcommand{\cL}{\mathcal{L}} %
\newcommand{\Ls}{L} %

\let\oldparagraph\paragraph
\renewcommand{\paragraph}[1]{\oldparagraph{#1.}}

\newtheorem{definition}{Definition}

\usepackage{tikz}
\usetikzlibrary{arrows.meta}
\usetikzlibrary{calc, babel, decorations.text}
\usetikzlibrary{positioning}
\usepgflibrary{shapes.arrows}
\usetikzlibrary{patterns}
\usetikzlibrary{decorations.text}    
\usetikzlibrary{intersections}
\usetikzlibrary{fit}
\usetikzlibrary{matrix}

\usetikzlibrary{shadows,spy,shapes.symbols, overlay-beamer-styles}

\makeatletter
\tikzset{arc style/.initial={}}
\pgfdeclareshape{half circle}{
    \inheritsavedanchors[from=circle]
    \inheritanchorborder[from=circle]

    \inheritanchor[from=circle]{center}
    \inheritanchor[from=circle]{south}
    \inheritanchor[from=circle]{west}
    \inheritanchor[from=circle]{north}
    \inheritanchor[from=circle]{east}

    \inheritbackgroundpath[from=circle]

    \beforebackgroundpath{
        \pgfkeys{/tikz/arc style/.get=\tmp}
        \expandafter\tikzset\expandafter{\tmp}
        \tikz@options

        \radius \pgf@xa=\pgf@x
        \centerpoint \pgf@xb=\pgf@x \pgf@yb=\pgf@y

        \advance\pgf@yb by \pgf@xa
        \pgfpathmoveto{\pgfpoint{\pgf@xb}{\pgf@yb}}
        \pgfpatharc{90}{-90}{\pgf@xa}

        \pgfusepath{fill}
    }
}
\makeatother

\newcommand{\decisionDashed}[7][]{
    \draw[thick, dashed] ([shift=(-1*#6:#5)]#3,#4) arc (-1*#6:#6:#5);
    \draw[] (#3,#4) -- ++(-1*#6:#5) (#3, #4) -- ++(#6:#5);
    \node(regulator)[circle, fill=black, label={[font=\footnotesize, align=center, rounded corners, fill=white, fill opacity=0.7, text opacity=1]#2}, pin={[pin distance=#7]\ifx\empty#1\relax\else #1\fi}] at (#3, #4) {};
}

\newcommand{\decisionDashedSides}[8][]{
		\draw[thick] ([shift=(-1*#6:#5)]#3,#4) arc (-1*#6:#6:#5);
		\draw[dashed] (#3,#4) -- ++(-1*#6:#5) (#3, #4) -- ++(#6:#5);
		\node [circle, fill=black, label={[name=n, font=\footnotesize, align=center, rounded corners, fill=white, fill opacity=0.7, text opacity=1]#8:#2}, pin={[pin distance=#7]\ifx\empty#1\relax\else #1\fi}] at (#3, #4) {};
		
}

\newcommand{\decisionDashedSidesUnder}[9][]{

        \def\col{\ifx#2B blue\else red\fi}
		\draw[thick] ([shift=(-1*#6:#5)]#3,#4) arc (-1*#6:#6:#5);
		\draw[dashed] (#3,#4) -- ++(-1*#6:#5) (#3, #4) -- ++(#6:#5);
		\node [circle, fill=\col, label={[name=n, font=\footnotesize, align=center, rounded corners, fill=white, fill opacity=0.7, text opacity=1, text=\col, font=\bfseries]#8:#2}, pin={[pin distance=#7, color=\col]below:\ifx\empty#1\relax\else #1\fi}] at (#3, #4) {};
		\node #9 at (n){};
}

\newcommand{\decisionDashedSidesOver}[9][]{
    	\def\col{\ifx#2B blue\else red\fi}
		\draw[thick] ([shift=(-1*#6:#5)]#3,#4) arc (-1*#6:#6:#5);
		\draw[dashed] (#3,#4) -- ++(-1*#6:#5) (#3, #4) -- ++(#6:#5);
		\node [every pin/.append style={color=\col}, circle, fill=\col, label={[name=n, font=\footnotesize, align=center, rounded corners, fill=white, fill opacity=0.7, text opacity=1, text=\col, font=\bfseries]#8:#2}] at (#3, #4) {};
		\node #9 at (n){};
}

\newcommand{\decisionUnder}[9][]{
    \draw[thick] ([shift=(-1*#6:#5)]#3,#4) arc (-1*#6:#6:#5);
    \draw[] (#3,#4) -- ++(-1*#6:#5) (#3, #4) -- ++(#6:#5);
    \node(regulator)[circle, fill=black, label={[name=n, font=\footnotesize, align=center, rounded corners, fill=white, fill opacity=0.7, text opacity=1]#2}, pin={[pin distance=#7]below:\ifx\empty#1\relax\else #1\fi}] at (#3, #4) {};
    \node[fill=white, fill opacity=0.7, text opacity=1] at (#3, #4 - #7 + 0.3) {\scriptsize #8}
    \node #9 at (n){};
}

\newcommand{\decisionFirst}[8][]{
    \def\col{\ifx#2Bblue\else red\fi}
    \draw[thick] ([shift=(-1*#6:#5)]#3,#4) arc (-1*#6:#6:#5);
    \draw[] (#3,#4) -- ++(-1*#6:#5) (#3, #4) -- ++(#6:#5);
    \node #8 at (#3, #4) {};
    \node  [circle, fill=\col, label={[font=\footnotesize, align=center, rounded corners, fill=white, fill opacity=0.7, text opacity=1, text=\col, font=\bfseries]#2}, pin={[pin distance=#7]\ifx\empty#1\relax\else #1\fi}] at (#3, #4) {};
}

\newcommand{\decision}[8][]{
    \def\col{\ifx#2Bblue\else red\fi}
    \draw[thick] ([shift=(-1*#6:#5)]#3,#4) arc (-1*#6:#6:#5);
    \draw[] (#3,#4) -- ++(-1*#6:#5) (#3, #4) -- ++(#6:#5);
    \node #8 at (#3+#5/4, #4) {};
    \node  [circle, fill=\col, label={[font=\footnotesize, align=center, rounded corners, fill=white, fill opacity=0.7, text opacity=1, text=\col, font=\bfseries]#2}, pin={[pin distance=#7]\ifx\empty#1\relax\else #1\fi}] at (#3, #4) {};
}

\tikzset{
    partial ellipse/.style args={#1:#2:#3}{
            insert path={+ (#1:#3) arc (#1:#2:#3)}
        }
}

\pgfdeclarepatternformonly{super ultra thick north east lines}{\pgfqpoint{-2pt}{-2pt}}{\pgfqpoint{10pt}{10pt}}{\pgfqpoint{10pt}{10pt}}%
{
  \pgfsetlinewidth{2pt}
  \pgfpathmoveto{\pgfqpoint{-1pt}{-1pt}}
  \pgfpathlineto{\pgfqpoint{9pt}{9pt}}
  \pgfusepath{stroke}
}

\usetikzlibrary{external}
\tikzexternaldisable %

\title{Regulation Games for Trustworthy Machine Learning}

\author[1]{Mohammad Yaghini*}
\author[1]{Patty Liu}
\author[1]{Franziska Boenisch}
\author[1]{Nicolas Papernot}

\affil[1]{University of Toronto and Vector Institute}

\begin{document}

{
    \maketitle
    \def\thefootnote{*}\footnotetext{Corresponding author: \texttt{mohammad.yaghini@mail.utoronto.ca}}
}

\begin{abstract}
Existing work on trustworthy machine learning (ML) often concentrates on individual aspects of trust, such as fairness or privacy. \stephan{Can you think of a better transition here? It is not clear at first sight how these are related} Additionally, many techniques overlook the distinction between those who train ML models and those responsible for assessing their trustworthiness. To address these issues, we propose a framework that views trustworthy ML as a multi-objective multi-agent optimization problem. This naturally lends itself to a game-theoretic formulation we call regulation games\stephan{Maybe emph this?}. We illustrate a particular game instance, the \specg in which we model the relationship between an ML model builder and fairness and privacy regulators\stephan{Maybe say something about the fact that this neatly combines the previously stated desiderata into an end-to-end solution?}. Regulators wish to design penalties that enforce compliance with their specification, but do not want to discourage builders from participation. Seeking such socially optimal (\ie, efficient for all agents) solutions to the game, we introduce \pp. This novel equilibrium search algorithm ensures that agents remain on the Pareto frontier of their objectives and avoids the inefficiencies of other equilibria. Simulating \specg through \pp can provide policy guidance for ML Regulation. For instance, we show that for a gender classification application, regulators can enforce a differential privacy budget that is on average 4.0 lower if they take the initiative to specify their desired guarantee first.
  
  \end{abstract}
  
\section{Introduction}
\label{sec:intro}

In most applications, it is impossible to eliminate risks of deploying machine learning (ML) models. This is because responsible deployment of ML models involves challenges beyond optimizing for model utility, and includes, among others, algorithmic fairness~\citep{pedreshi_discrimination-aware_2008, calders2010three}, privacy~\citep{blum_practical_2005, abadi2016deep}, robustness~\citep{szegedy2013intriguing}, and interpretability~\citep{linardatos_explainable_2020}. These objectives invariably present various trade-offs with one another. \citet{chang_privacy_2021, vinith} discuss the fairness-privacy trade-off, \citet{kifer_no_2011} considers utility-privacy \stephan{Shameless plug: if you want, you can cite my uncertainty-privacy paper :D. Would also fit well in the first sentence in the related work section}. The fairness-utility trade-off is also widely studied~\citep{wick_unlocking_2019}.
Recently, works such as~\citet{esipova_disparate_2023, impartiality} considered the 3-way trade-off between fairness, privacy and utility.

Regardless of the method, prior work implicitly assumes that a \textit{single} entity is in charge of implementing the different objectives. 
Unfortunately,
regarding trust in ML purely from a single-agent lens runs the risk of producing trade-off recommendations that are \textit{unrealizable} in practice.  This is because nowadays ML models 
are trained and audited by separate entities---each of which may pursue their own objectives. To carry out the aforementioned trade-off recommendations, it would require the agents to align their objectives and take coordinated action. \stephan{Maybe explicitly state that (or even better why) this is bad/inefficient or maybe even impossible sometimes?}

Given the multitude of agents and objectives involved, we argue that achieving trust in ML is inherently a multi-objective multi-agent problem~\citep{radulescu_multi-objective_2020}. Since Game Theory is the natural tool to model and analyze the interactions between different agents, we initiate the first study on multi-agent trustworthy ML within a novel class of games we call the \textit{ML Regulation Games}.

We present \specg, a novel game theoretic formulation of ML regulation in the form of a repeated game between regulators and the model builder. 
We illustrate the application of \specg with regulators who produce specifications for fairness and privacy guarantees \stephan{Is the framework more general? Does it allow other notions of trustworthiness to be included into the regulation game? If so, maybe hint at that}. The primary concern of the model builder is utility, which means builders have incentive to exceed the specifications and produce untrustworthy models. In this case, regulators step in and enforce compliance by designing monetary penalties. If a penalty is too low, the builder may absorb it as ``cost-of-doing-business,'' rendering it ineffective. On the other hand, if the penalty is too high, it can dissuade builders from participation. Ideally,  we seek penalties that are proportional to the violation from the specification---thus discouraging untrustworthy models.

\specg can be simulated via standard algorithms such as Best-Response Play~\cite{fudenberg_game_1991} to search for its Nash equilibria.
However, defined within the context of ML regulation, we observe a structure particular to \specg, that allows us to search for Correlated equilibria instead---a super-set of Nash equilibria known to be more efficient to calculate~\citep{arora_multiplicative_2012, nisan_algorithmic_2007}:
\begin{quote}
\textit{The underlying ML task and the corresponding data generating phenomenon is common between the regulators and the builder.}
\end{quote}
For instance, if the ML task is gender estimation, then the regulators and the builder both sample data from the same phenomenon (gendered human faces) even if their datasets are different. This fact allows agents to self-coordinate and correlate their strategies. We take advantage of this fact to present \pp, a novel equilibrium search algorithm for \specg that allows shared access to the Pareto frontier of agent objective losses.

We prove that \pp produces correlated equilibria. \stephan{Can you spell out more clearly why a \textit{correlated} equilibrium is desired? Does correlated just mean "good for both"? Maybe this is obvious for people with the approproate game theory background}
Simulating these equilibria enables us to engage in empirical incentive design where our goal remains designing penalties that are effective in enforcing compliance with the specification and avoid ``bad'' equilibria such as the aforementioned price-of-doing-business.

Our formulation of \specg and \pp are general in the sense that they do not assume a particular private learning framework (e.g. DP-SGD~\citep{abadi2016deep} or PATE~\citep{papernot2016semi}), or a specific notion of fairness like demographic parity \stephan{Citation?}. Furthermore, they can be extended to enforce regulations for objectives beyond fairness and privacy. We showcase this general applicability by instantiating \specg with two learning algorithms that are both differentially-private and fairness-aware~\citep{impartiality, esipova_disparate_2023}.
\stephan{Should this be its own paragraph?} In our empirical evaluations, we highlight the insights that simulating \specg brings towards designing more effective ML regulation. For instance, we show that for a gender classification application, regulators can enforce a differential privacy budget that is on average 4.0 lower if they take the initiative to specify their desired guarantee first. \stephan{Seems a bit short. Are there any other cool results from the empirical results that are worth being highlighted?}

In summary, we make the following contributions: 

\begin{itemize}
\item We introduce \specg, a general class of games to aid ML regulators in the specification and enforcement of their policies.

\item We propose \pp, a novel simulation algorithm that takes advantage of the fact that agents share a common data distribution to allow agents to coordinate using self-calculated \pfs.  
\pp allows efficient simulation of \specg to recover equilibria. Analysis of these equilibria allows us to tackle questions of \textit{empirical incentive design} to push for effective, realizable and proportional ML regulation.
\item Through empirical simulations of \specg, we provide policy guidance for ML Regulation and validate our design choices. Notably, we highlight the suboptimality of studying trustworthy ML in a single-agent framework.

\end{itemize}

\section{Background}
\label{sec:background}
\paragraph{Problem Setting: Specification}
Consider an example scenario involving an application of ML to automated facial analysis---similar to models considered in~\citet{krishnan2020understanding}. This is for illustrative purposes; our framework is generally applicable. \stephan{Feels to me like the pas sentence should be a footnote.}
Facial analysis models perform a range of sensitive tasks, such as face recognition and prediction of sensitive attributes like gender or age.\footnote{The use of such models can, therefore, have severe ethical implications, motivating the need to optimize for their general trustworthiness, instead of focusing on one single aspect, such as their average utility. %
} \stephan{I think this sentence would feel more natural in the main body.}
It is clear that both fairness and privacy are relevant to the individuals that use such a model or contribute to its training data.
In practice, regulators, such as governmental agencies or ethics committees, are put in charge of formulating and enforcing privacy and fairness requirements on the individuals' behalf. This is why it behooves us to consider them as independent \textit{agents} in ML regulation. \stephan{Based on the start of the paragraph, I was expecting a more detailed formalism of the example setting. Is this added somewhere later? It also seems like all other paragraphs in this section have at least a little bit of formalism.}

\paragraph{Game Theory} A \textit{Stackelberg} competition $\mathcal{G} = {(\cA, \cS, \cL)}$ is a game defined by the set of agents $\cA$, their strategies $\cS$, and their loss functions $\cL$, where the players go in sequential order~\citep{fudenberg_game_1991}. In its simplest form with two agents, the \textit{leader} initiates the game. Having observed the leader's strategy, a \textit{follower} reacts to it. Analysis of Stackelberg competitions, involves solving a bi-level optimization problem where every agent is minimizing their loss subject to other agents doing the same. Solutions to the said problem recovers the \textit{Nash equilibria}.
Searching for Nash equilibria is NP-hard 
~\citep{ne_ppad}, which is why a super-set of them, known as \textit{Correlated equilibria} have seen increasing attention due to ease with which they can be found (for instance, using polynomial weights algorithm)~\citep{arora_multiplicative_2012, nisan_algorithmic_2007}.

In multi-objective optimization, and games in particular, we are interested in the \textit{Pareto efficiency}. A tuple of objective values are Pareto efficient if we cannot improve one of the values without making another worse off. More formally, given objectives parameterized by ML models, we have:
\begin{definition}[Pareto Efficiency]   
\label{def:pareto}
A model $\model \in \mathcal{W}$, where $\mathcal{W}$ is the space of all models, is \pe if there exists no $\model' \in \mathcal{W}$ such that \textbf{(a)} $\forall i \in \cL$ we have $\ell_i\left(\model'\right) \leq \ell_i(\model)$ where $\cL$ is the set of losses and $\ell_i \in \cL$ is the objective $i$'s loss; and that \textbf{(b)} for at least one loss $j \in \cL$ the inequality is strict $\ell_j\left(\model'\right) < \ell_j(\model)$.
\end{definition}

\paragraph{Privacy}
Differential Privacy (DP)~\citep{dwork_algorithmic_2013} is a mathematical framework that provides rigorous privacy guarantees to individuals
contributing their data for data analysis. A typical DP mechanism is to add \textit{controlled noise} to the analysis algorithm, making it difficult to identify individual contributions while still yielding useful statistical results.
Formally, $(\varepsilon, \delta)$-differential privacy is expressed as follows: \stephan{"Our work considers the $(\varepsilon, \delta)$-differential privacy setup, which is expressed as follows:"} %
\begin{definition}[$(\varepsilon, \delta)$-Differential Privacy]
\label{def:DP}
Let $\mathcal{M} \colon \mathcal{D}^* \rightarrow \mathcal{R}$ be a randomized algorithm that satisfies $(\varepsilon, \delta)$-DP with $\varepsilon \in \mathbb{R}_+$ and $\delta \in [0, 1]$ if for all neighboring datasets $D \sim D'$, \ie, datasets that differ in only one data point, and for all possible subsets $R \subseteq \mathcal{R}$ of the result space it must hold that
$\varepsilon \geq \log{\frac{\mathbb{P}\left[\mathcal{M}(D) \in R\right] - \delta}{\mathbb{P}\left[\mathcal{M}(D') \in R\right]}}.$ \stephan{This definition reads a bit cumbersome to me. Maybe split into two sentences? "Let $\mathcal{M} \colon \mathcal{D}^* \rightarrow \mathcal{R}$ be a randomized algorithm. We say that $\mathcal{M}$ satisfies $(\varepsilon, \delta)$-DP with $\varepsilon \in \mathbb{R}_+$ and $\delta \in [0, 1]$ if for all neighboring datasets $D \sim D'$, \ie, datasets that differ in only one data point, and for all possible subsets $R \subseteq \mathcal{R}$ of the result space it must hold that
$\varepsilon \geq \log{\frac{\mathbb{P}\left[\mathcal{M}(D) \in R\right] - \delta}{\mathbb{P}\left[\mathcal{M}(D') \in R\right]}}.$"}
\end{definition}

DP, by definition, protects against worst-case failures of privacy~\citep{dwork_algorithmic_2013, mironov_renyi_2017}. These failures are rare events and difficult to detect. As a result, sample-efficient estimation of DP guarantees is nearly impossible~\citep{gilbert_property_2018}. In~\Cref{sec:specification-game}, we circumvent this issue by training models on a separate data set from the same data distribution which provides us with estimations for the privacy budget required to perform the underlying ML task. \stephan{I don't think I quite understand from this explanation what is going on here.}

\paragraph{Fairness} There are abundant notions of fairness and corresponding measures in the literature~\cite{fairml}. 
The choice of fairness measure is largely task-dependent and at the behest of the regulators. Hence, our regulation game framework does not make any assumptions on the applied metric. The fairness evaluation process takes as input a fairness metric $\Gamma_{\text{fair}} (\model, D): \mathcal{W} \times \mathcal{X} \mapsto \posReals$ chosen by the regulator, the model $\model \in \mathcal{W}$, and an adequate evaluation dataset $\Deval \in \mathcal{X}$, 
$\Deval \sim \mathcal{D}$, where $\mathcal{D}$ is the task's data distribution.  The evaluation process then outputs $\hgmm_\model$ as an empirical estimate of the model's fairness violation. In~\Cref{sec:results}, we instantiate concrete ML algorithms with their stated fairness measures which we discuss in detail in~\Cref{sec:fairness}.

\section{\protect\specg}
\label{sec:specification-game}

We introduce \specg, an ML regulation game that captures the interactions between 
three agents involved in the life-cycle of an ML model~\citep{tomsett2018interpretable}: a \emph{model builder} who is in charge of producing the model, and two regulators who are in charge of fairness and privacy of the resulting model, respectively. We note that our framework is general and can accommodate other objectives, as long as they are measurable with a loss function. We assume the model builder seeks to create the most accurate model. 
The regulators formulate the requirements and monitor the model for potential violations of their objectives based on recent regulations~\citep{aiact2021EU,canadaPersonalInformationProtection2019}.
The \textit{fairness regulator} assesses the resulting model for potential discrimination using fairness metrics~\citep{beutel2019putting}, whereas the \textit{privacy regulator} seeks to ensure that strong-enough guarantees exist to protect the privacy of the training data used for the model.
We assume that regulators are able to give penalties\footnote{As is customary in the economics literature, the penalties need not be monetary. It is sufficient that they present a viable risk. This, for instance, may take the form of expected lost revenue due to a watch-dog (or NGO) reporting on fairness and/or privacy violations of an ML-as-a-Service platform.} for violations of their respective objective which they formulate as a loss function.

\begingroup
\setlength{\columnsep}{3.2mm}%
\setlength{\intextsep}{2mm}%
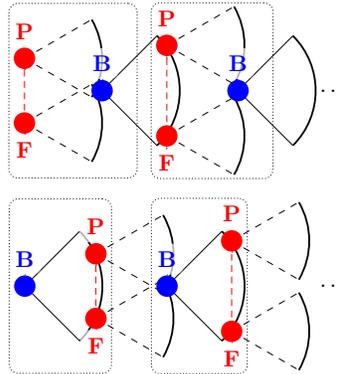
\begin{wrapfigure}{r}{0.3\textwidth}
    \centering
    \resizebox{\linewidth}{!}{
            
\begin{tikzpicture}[]
    \def\dx{1.2}

    \decisionDashedSidesUnder[]{P}{0}{0.5}{\dx}{30}{0}{above}{(n1)};
	\decisionDashedSidesUnder[]{F}{0}{-0.5}{\dx}{30}{0}{below}{(n2)};	
	\decision[]{B}{\dx}{0}{\dx}{45}{0}{(n3)};
	\draw[densely dashed, draw=red] (0, 0.5) -- (0, -0.5);
	\node [draw, rounded corners, fit=(n1)(n2)(n3), densely dotted, minimum height=2.7cm]{};
	
	\decisionDashedSidesUnder[]{P}{2*\dx-0.2}{0.7}{\dx}{30}{1}{above}{(n4)};
	\decisionDashedSidesUnder[]{F}{2*\dx-0.2}{-0.7}{\dx}{30}{1}{below}{(n5)};	
    \decision[]{B}{2.75*\dx}{0}{\dx}{45}{0}{(n6)};
    \node [draw, rounded corners, fit=(n4)(n5)(n6), densely dotted, minimum height=2.7cm]{};
	\draw[densely dashed, draw=red] (2*\dx-0.2, 0.5) -- (2*\dx-0.2, -0.5);
	\node at (4*\dx, 0) {$\dots$};
    \node at (0, -1.5) {};
\end{tikzpicture}
    }
    \resizebox{\linewidth}{!}{
        \begin{tikzpicture}[]
    \def\dx{1.2}

    \decisionFirst[]{B}{0}{0}{\dx}{45}{0}{(n1)};
    \decisionDashedSidesUnder[]{P}{\dx-0.1}{0.5}{\dx}{30}{0}{above}{(n2)};
	\decisionDashedSidesUnder[]{F}{\dx-0.1}{-0.5}{\dx}{30}{0}{below}{(n3)};	
	\draw[densely dashed, draw=red] (\dx-0.1, 0.5) -- (\dx-0.1, -0.5);
	\node [draw, rounded corners, fit=(n1)(n2)(n3), densely dotted, minimum height=2.7cm]{};

    \decisionFirst[]{B}{2*\dx-0.2}{0}{\dx}{45}{0}{(n6)};	
	\decisionDashedSidesUnder[]{P}{2.75*\dx-0.1}{0.7}{\dx}{30}{0}{above}{(n4)};
	\decisionDashedSidesUnder[]{F}{2.75*\dx-0.1}{-0.7}{\dx}{30}{0}{below}{(n5)};	
    \node [draw, rounded corners, fit=(n4)(n5)(n6), densely dotted, minimum height=2.7cm]{};
	\draw[densely dashed, draw=red] (2.75*\dx-0.1, 0.5) -- (2.75*\dx-0.1, -0.5);
	
	\node at (4*\dx, 0) {$\dots$};
\end{tikzpicture}
    }
    \caption{\textbf{Repeated \specg between model \underline{B}uilder, and \underline{P}rivacy and \underline{F}airness regulators}---\textit{(top) }regulators lead, \textit{(bottom)} builder leads. 
    }
    \label{fig:game}
\end{wrapfigure}
Depending on whether regulators announce concrete specifications first, or if the model builder produces a model first with fairness and privacy guarantees of its choosing, we would have a game that is either \textit{regulator-led}, or \textit{builder-led}. This sequential order of interactions lends itself naturally to a Stackelberg competition (see~\Cref{sec:background}).
As agents interact with each other, the game is repeated:~\Cref{fig:game} 
shows both a \textit{regulator-led} (top) and a \textit{builder-led} market (bottom). Since analysis of both settings is similar, without loss of generality (W.L.O.G), unless otherwise stated, we will assume a regulator-led market. 
In~\Cref{sec:results}, we will discuss the impact of moving first.

\dtodo{
We assume the regulators hold necessary information about the task at hand in the form of a Pareto Frontier (\pf)\footnote{It has been shown that this does not require access to private data since access to data from the same domain is sufficient to calculate the \pf~\cite{impartiality}.}\todo{this is important - but probably not right now - that said, don't leave this to a footnote} which they use to choose fairness and privacy requirements that taken together with the resulting accuracy loss are \textit{Pareto efficient}: improving one objective would necessarily come at the cost of another (see~\Cref{def:pareto}). This choice departs from the classical non-cooperative game formulations but we argue it is appropriate given that regulators do not seek to punish model builders for creating accurate models given that a well-generalized model is necessary for strong privacy~\citep{li_large_2022} and fairness guarantees~\citep{disp-vuln}.
}

Every \specg is parameterized by a specification $\bs_\reg$ which is a tuple of guarantee levels that regulators deem to be safe; for instance $\bs_\reg = (\gmm, \eps)$ where $\gmm$ is the maximum tolerable fairness violation, and $\eps$ is the maximum privacy budget. The choice of fairness metric is task-dependent and the choice for privacy metric depends on the privacy notion that the regulators find best protects user privacy.  The model builder is in charge of implementing an ML task (e.g., gender estimation) by producing a model~$\model \in \mathcal{W}$ where $\mathcal{W}$ is the space of all possible models which is then passed onto regulators to audit, and if necessary, assign an appropriate penalty.

\subsection{Game Formalization}
\label{sec:game-formalization}

Formally the \specg $\mathcal{G}$ is an infinitely-repeated Stackelberg game. Its stage game $\mathcal{G}_\text{stage} = (\cA, \cS, \cL)$ is repeated across time $t \in \{0, 1, \dots\}$ as shown in~\Cref{fig:game} where each stage in marked with dotted windows. $\cA = \{\build, \fair, \priv\}$ is the set of agents.
The strategy space of the stage game is $\cS = \{(\bs_\fair, \bs_\priv, \bs_\build)\}$ and $\cL = \{(\ell_\build, \ell_\fair, \ell_\priv)\}$ represent agent losses. The complete game is thus defined as the Cartesian product of the stage game: $\mathcal{G} = \mathcal{G}_\text{stage}^n$ where $n \rightarrow \infty$ (i.e., infinitely repeated). To analyze $\mathcal{G}$ we are interested in the overall \textit{discounted} loss of agent $i \in \cA$ defined as $\bar{\ell}_i=\sum_{t=0}^{\infty}c^t \ell_i^{(t)}$.
$c$ is known as the \textit{discounting factor} and represents the fact that agents care about their loss in the near-term more than in the long run~\citep{mas}.

We can formalize agent $i$'s strategy space by their observable actions: the model $\model$ released by the builder, or the penalties $L_\fair$, $L_\priv$ announced by the regulators. Alternatively, we can formulate the agent strategies in terms of their \textit{latent} acceptable trustworthy parameters $\bs_i$. That is, we assume that instead of choosing models and penalties, agents specify their trustworthy parameters $\bs_i$ in response to each other. We note that in this case, $\bs_i$ are not directly observable by other agents, rather we have $\model = f_1(\bs_\build)$, $L_\fair = f_2(\bs_\fair)$ and $L_\priv = f_3(\bs_\priv)$ provided there exists mappings $f_1, f_2$ and $f_3$. 
Under the assumption of a shared Pareto frontier, these maps do exist \stephan{Can they only exist there? If so, why?}. We discuss this assumption in~\Cref{sec:assumption} and take advantage of this formulation in~\Cref{sec:incentive-design}.
The latent strategy representation simplifies the game by taking the agent interactions from the space of penalties and high-dimensional models to the much smaller space of trustworthy parameters. Next, we present each agent and formalize their loss and strategy.

\paragraph{Fairness Regulator}
Assuming regulators lead, at the first stage of \specg, the fairness regulator's strategy is to specify the maximum tolerable fairness violation $\gmm \in \posReals$. Next, the builder creates the model $\model$. At the end of the first stage, where losses are measured, the regulator's loss is
\begin{equation}
    \label{eq:fairness_loss}
    \ell_\fair(\gmm; \model) := \max \{0, \hgmm_\model - \gmm\},
\end{equation}
where $\hgmm_\model$ is the regulator's evaluation of model $\model$ fairness violation (see~\Cref{sec:background}). \stephan{Hinge-type loss makes sense to me but maybe it would be good to briefly explain why this is a good choice.}
In the subsequent stages of the game, the regulator strategy is to announce a penalty (\eg, a monetary fine) $\Ls_\fair \in \cL_\fair$ for violating its specification where $\cL_\fair = \posReals$ is the strategy space of the regulator. 
We require that the fine be proportional to the excessive fairness violation ($\hgmm_\model -\gamma$) and choose the linear form $\Ls_\fair(\model) = \cfair \ell_\fair(\gmm; \model)$ \stephan{Any reason why this is linear? Are more complex functions possible?}. This formulation also follows the the common ``expected utility hypothesis'' in economics~\citep{roberts_measurement_1984}. If $\cfair$ is too small, the regulator's strategy may not be effective in persuading the model builder to create a fairer model and thus lower $\hgmm_\model$. We discuss how to chose $\cfair$ in~\Cref{sec:incentive-design}.

\paragraph{Privacy Regulator}
Similar to the fairness regulator, the privacy regulator's only requirement for the privacy definition is that its violations be measurable. However, unlike fairness, privacy has a general and well-established computational notion of privacy in the form of differential privacy~\citep{dwork_dp} (DP). Therefore, we assume that the privacy regulator specifies its requirement within the framework of $(\varepsilon, \delta)$-DP
(see~\Cref{def:DP}).

The privacy regulator needs to estimate the privacy of the builder's model. In our formulation we abstract away the particular technique used for privacy estimation.  We note that the technique used bears importance in the simulation of the game, but we defer that discussion until~\Cref{sec:pareto-play} where we discuss our simulator for \specg.
Concretely, we treat privacy auditing as a function that, at a fixed $\delta$, outputs an estimate $\heps_\model \in \posReals$ of the privacy parameter $\eps_\model$ used by the model builder. 
Based on this estimate and the privacy parameter $\eps \in \posReals$ required by the privacy regulator, we formulate the privacy loss as: 
\begin{equation}
    \label{eq:privacy_loss}
     \ell_\priv (\model) :=  \max\{0, \heps_\model - \eps\}.
\end{equation}
Similar to the fairness regulator, the privacy regulator strategy is to choose a penalty $\Ls_\priv \in \cL_\priv$ where $\cL_\priv = \posReals$ is its strategy space. We again choose the penalty to be proportional to the excessive privacy leakage $(\heps_\model - \eps)$. Therefore, $\Ls_\priv(\model) = \cpriv \ell_\priv (\model)$.

\endgroup

\paragraph{Model Builder}
The model builder produces the model~$\model \in \cS_\build$ where $\cS_\build = \mathcal{W}$ is the space of all possible models. While builder's primary focus is achieving high model utility (\eg, accuracy), they are aware that the released model will have to face regulatory audits.
In particular, the builder knows that fairness or privacy violations from $\bs_\reg = (\gmm, \eps)$ can lead to penalties from the regulators.
The model builder's loss $\ell_\build \in \cL_\build$ is its model error $\err_{build}(\model)$ on a test dataset $\testdataset \sim \mathcal{D}$, where $\mathcal{D}$ is the data distribution, plus the penalties incurred from privacy and fairness regulators ($\Ls_\priv$ and $\Ls_\fair$) that are scaled back to its error space using $\lambda_\priv$ and $\lambda_\fair$, respectively:
\begin{align}
   \ell_\build(\model) &= \err_\build(\model; D_\text{test})+ \buildPriv \Ls_\priv(\model) + \buildFair \Ls_\fair(\model).
   \label{eq:model_builder_loss}    
\end{align}

Finally, the builder's strategy is to release the model that minimizes its loss: $\model = \argmin_{\model} \ell_\build(\model)$. \stephan{Maybe worth reiterating that this is the game loss for the model builder (at least according to my understanding), and not the actual training loss of the model builder.}

\paragraph{An interpretation of builder penalty scalars $\lambda_\fair$ and $\lambda_\priv$}
\label{par:lambda-interpretation}
Assume two models $\model_1$ and $\model_2$ produce the same $\ell_\build$ and $\Ls_\fair$ but different $\err_\build$ and $\Ls_\priv$. We have $\buildPriv := - {\Delta\err_\build}/{\Delta\Ls_\priv}$ where $\Delta\{.\}$ denotes the difference between $\model_1$ and $\model_2$. In other words, \textit{$\buildPriv$ is the improvement in model builder's error given a unit increase in privacy penalty (which is proportional to a unit increase in privacy leakage).} 
A similar interpretation holds for $\buildFair$. We will use this interpretation in~\Cref{sec:incentive-design}. We note that $\buildFair$ and $\buildPriv$ are known only to the builder. In~\Cref{sec:lambda-calib}, we will discuss how the regulator can estimate them in practice.

\section{\pp: Best-Response Play on the Pareto Frontier}
\label{sec:pareto-play}

Although Nash equilibria are optimal w.r.t. single-agent deviations, they are often not societally optimal (see~\Cref{def:pareto}). For instance, seeking NEs can provide `solutions' where both the model builder and a regulator's losses can be improved simultaneously which is not a desirable outcome for ML regulation. Furthermore, the \specg ~described in \Cref{sec:specification-game} cannot be easily simulated directly due to challenges in forming the agents' loss functions, most notably, because privacy violations of a trained model is difficult to estimate in a black-box fashion without access to its training procedure~\citep{gilbert_property_2018}. 

In~\Cref{sec:simulating-paretoplay} we introduce \pp to address these problems by assuming agents share as common-knowledge a pre-calculated \pf between privacy, fairness, and model utility. We discuss the validity of this assumption in~\Cref{sec:assumption}. In~\Cref{sec:incentive-design} we consider incentive design for $C_\fair$ and $C_\priv$ to ensure regulation is effective. We discuss how regulators can estimate builder's penalty scalars $\lambda_\fair$ and $\lambda_\priv$
in~\Cref{sec:lambda-calib}.

\subsection{Simulating \specg with \pp}
\label{sec:simulating-paretoplay}

The game starts by distributing an initial \pf between all agents. The \pf is formed by training multiple instances of the chosen ML models in $R=\{\model (\bs) \mid \model \in \mathcal{W}\}$ before the game using different guarantee levels $\bs := (\gmm, \eps)$ and then calculating the Pareto frontier $P = \operatorname{PF}(R)$ where $\operatorname{PF}: \posReals \times \posReals \mapsto \posReals \times \posReals \times \posReals$ 
is a map from trustworthy parameters to a tuple of achieved fairness, privacy and builder losses.

\begin{algorithm}[thb]
\caption{\textbf{\pp}: Regulator-led}
\label{alg:paretoplay}
\begin{algorithmic}[1]
    \Input{
    Initial \pf input $R^{(0)}$,
    total number of game rounds $T$,
    agents \{$A_\build$, $A_\reg \mid \reg \in \{\fair, \priv\}\}$,
    step size $\eta$
    }

        \For{$t \in \{0, 1, \dots, T\}$}
            \State $P \gets \operatorname{PF}(R^{(t)})$ \Comment{Estimate the \pf} \label{line:pareto-calc}
            \If {t = 0} \Comment{First round of the game}
                \State $\bs^0 \gets \Call{ChooseSpec}{P, \{A_\fair, A_\priv\}}$ 
            \ElsIf{$t \mod 3 \neq 1$} \Comment{Regulators moves}
                \State ${C^{(t)}_\reg \gets \Call{ChoosePenaltyScale}{\bs^{(t-1)}}}$
                \State ${\bs^{(t+1)} \gets \bs^{(t)} - \eta \left<\boldsymbol{e}_\reg, \nabla_s \Ls_\reg^P(\bs^{(t)}; C^{(t)}_\reg)\right>}$ \Comment{\reg~only adjusts its own parameter} \label{line:reg-update}
            \Else \Comment{Builder move}
            \State $\bs^{(t+1)} \gets \bs^{(t)} - \eta \nabla_s \ell^{P}_\build(\bs^{(t)})$ 
        \EndIf
        \State $R^{(t+1)} \gets R^{(t)} \cup \{\Call{Calibrate}{\model, \bs^{(t+1)}}\}$
        \State $\eta \gets c_i \cdot \eta$ \Comment{$A_i$ discounts its payoff by $c_i$} (decay factor) \label{line:decay}
        \EndFor
        \State {\bf Output} $\bs^{(T)}$ 
\end{algorithmic}
\end{algorithm}

\paragraph{Approximation and calibration} In \pp, we estimate all agent losses on the \pf $P$ (hence the $\ell^{P}_*$ in~\Cref{alg:paretoplay}). Our estimation involves a linear interpolation on $P$.
Interpolation may lead to estimation errors, as the estimated next parameters $\bs^{(t+1)}$ may, in fact, not be on the \pf. Given that the \pf is shared, this can lead to compounding errors, and non-convergent behavior. We avoid this by including a \textit{calibration} step at the end of each round. In $\Call{Calibrate}{}$ in~\Cref{alg:paretoplay}, we train a new model using the chosen parameter $\bs^{(t+1)}$, and add new objective loss results to the prior result set $R$. A new, potentially improved \pf is recalculated next round with the new results (\cref{line:pareto-calc}). Next, we will show that \pp induces a correlated equilibrium.

\paragraph{ParetoPlay induces correlated equilibria} 
Sharing the \pf has important implications for the equilibrium search: the Pareto frontier gives a signal to every player what to play (similar to how a stop-light allows drivers to coordinate when to pass an intersection).
This is known as a \textit{correlation device}. If playing according to the signal is a best response for every player, we can recover a correlated equilibrium.
\begin{theorem}
\label{thm:cne}
\pp recovers a Correlated Nash Equilibrium of the \specg. 
\end{theorem}
We defer the proof to~\Cref{sec:proofs}. We also provide a proof sketch for the convergence of \pp in~\Cref{thm:convergence}.
We discussed how \pp ensures convergence to a correlated equilibrium given access to a common-knowledge \pf. It is natural to ask if the assumption of a common-knowledge \pf requires \textit{information equality} between the model builder and the regulators. More concretely, should agents have the exact \pf for \pp to be viable?  In~\Cref{sec:assumption}, we argue that the answer is negative.

\subsection{On the Validity of the Shared Pareto Frontier}
\label{sec:assumption}

 We show that it is not necessary to assume that the \pfs of the builder and the regulators are the same. Rather, it is enough to assume that the datasets they are calculated on are from the same data-generating distribution. Concretely, we show that, the problem of finding the \pf for each agent can be written as a multi-objective optimization problem, the solution to which reduces to empirical risk minimization in ML. We conclude that the assumption of the shared Pareto frontier between agents is akin to the standard assumption of IID-ness (independent and identically-distributed data) in ML.

\paragraph{Deriving the \pf via scalarization}
There exist standard techniques to recover the Pareto frontier  of a multi-objective optimization problem---which always exists for any feasible problem.  
\textit{Scalarization}~\cite[Section 4.7.4]{boyd_convex_2004} is such a technique that, provided each objective is convex, can recover all of the Pareto frontier; and if not, at least a part of it. 
For our problem, the objective loss of the scalarized problem is 
$
   \min_{\bs}\alpha_1 \ell_\build(\bs) + \alpha_2 \ell_\fair(\bs) + \alpha_3 \ell_\priv(\bs),
$
where $\alpha_1,\alpha_2,\alpha_3 \geq 0$ are free parameters, different choices for which will give us various points on the Pareto frontier. 
Implicit in the scalarized objective loss are two assumptions: a) the dataset used to optimize the loss, and b) dependency on model weights $\model$. Making these assumptions explicit allows us to write the Pareto frontier $PF_i$ calculated by agent $i$:
\begin{equation}
   PF_i = \{\argmin_{\bs}\; \min_\model \alpha_1 \ell_\build(\bs, \model; D_i) + \alpha_2 \ell_\fair(\bs, \model; D_i) + \alpha_3 \ell_\priv(\bs, \model; D_i) \mid \alpha_1, \alpha_2, \alpha_3 \in \posReals\},
   \label{eq:scalarization-with-model}
\end{equation}
where $PF_i$ is calculated over dataset $D_i$ by agent $i$. \stephan{The formalism above suggests that $D_i$ is shared across all agents, but I guess you said that these can be different (but from the same distribution)? is there a better way to formalize that?} Seen through an ML lens,~\Cref{eq:scalarization-with-model} closely resembles an empirical risk minimization (ERM) problem. We optimize model parameters $\model$ in the inner sub-problem and tune the hyper-parameters $\bs$ in the outer one.

Coming back to question of whether \pfs are similar for different agents, we argue that since the problem of finding in the \pf reduces to an ERM problem, despite $D_i$ not being the same, we expect that the \pfs would be similar provided that $D_i \sim \mathcal{D}$ where $\mathcal{D}$ is the data-generating distribution, and that each $D_i$ have enough samples.
In other words, \textit{the true correlation device in \pp is not so much the \pf, but the real-world phenomenon whose data is sampled by each agent.}

We conclude this section by noting that prior works supports our assumption as well. \citet[Section 5.1.4]{impartiality} empirically showed that the \pfs calculated on separate datasets but for the same task are quite similar. In~\Cref{sec:results} under~\textit{\nameref{par:assumption}}, we empirically evaluate the shared \pf assumption. We simulate a \specg using \pp where regulators and model builders use different datasets but for modeling the same task (gender estimation). \pp converges because all agents are modeling the same data-generating phenomenon (gendered humans).

\subsection{Empirical Incentive Design: How to Set Penalty Scalars $C_\fair$ and $C_\priv$}
\label{sec:incentive-design}

\pp ensures that the found equilibria are Pareto efficient. However, not all such equilibria are desirable to the regulators. The ``cost-of-doing-business'' scenario we noted in~\Cref{sec:intro} where regulators' penalties are not enough to effect a change in model builder's behavior is an example. If the regulators seek to escape such an equilibrium, they can adjust the penalty scalars $C_\fair$ and $C_\priv$ (hereon, $C_*$). If the regulators had no consideration for the model builder's loss, then they  should choose a very large $C_*$. However, this runs the risk of disincentivizing participation completely by increaseing the risk of insurmountable penalties for the builder. So how should we choose $C_*$?

Consider builder's step at $t+1$ as seen by the regulators,
\begin{align}
\bs^{(t+1)} 
          &= \bs^{(t)} - \eta \nabla_{\bs}\err_\build(\bs^{(t)}) 
                         - \eta \cpriv \nabla_{\bs} \buildPriv \ell_\priv(\bs^{(t)})
                         - \eta \cfair \nabla_{\bs} \buildFair \ell_\fair(\bs^{(t)}),\label{eq:builder-update}
\end{align}
where $\nabla_{\bs}$ denotes the gradient with respect to the trustworthy parameters $\bs$.
Note the penalty scalars. $\lambda_\fair, \lambda_\priv$  (hereon, $\lambda_*$) are private information to the builder: a company is not incentivized to disclose how much a governmental penalty would be affecting its decisions.

Given the incomplete information of the regulator about the loss of the model builder, an exact answer is not possible. But as~\Cref{alg:paretoplay} shows regulators do not need to know $\lambda_*$ to impact the builder's loss and force it to change strategy. This is because the penalty scalars $C_*$ chosen by the regulators can create the same effect.  We can leverage scalarization from~\Cref{sec:assumption} to reason about and provide guidance on how to choose $C_*$.
 Consider the scalarization problem:
 \begin{equation}
 \label{eq:eq_incentive}
      \min_{\bs}\alpha_1 \err_\build(\bs) + \alpha_2 \ell_\fair(\bs) + \alpha_3 \ell_\priv(\bs).
 \end{equation}
Note that for ease of presentations, we are using the simplified notation of $\ell_i(\bs) := \ell_i(\bs, \model^\star, D_\reg)$ which implicitly assumes an optimal model $\model^\star$ is trained by the regulator(s) on their own dataset $D_\reg$ in the inner optimization of~\Cref{eq:scalarization-with-model}.
 As before, 
different choices for $\alpha_i$'s will give us various points on the Pareto frontier. 
Under the assumption of convexity then, all the steps in ParetoPlay are minimizers of the scalarized problem. Matching~\Cref{eq:eq_incentive} \stephan{What is this symbol doing?} with~\Cref{eq:builder-update} shows that $\alpha_1 \equiv 1$, $\alpha_2 \equiv \lambda_\fair C_\fair$ and $\alpha_3 \equiv \lambda_\priv C_\priv$. 

Using \textit{\nameref{par:lambda-interpretation}} from~\Cref{sec:specification-game}, we can see that $C_\fair \equiv 1/\lambda_\fair \equiv - \Delta \Ls_\fair/\Delta \err_\build$. In other words, \textit{$C_\fair$ should be chosen to offset any relative gains from excessive fairness violations.} A similar interpretation holds for~$C_\priv$. 

More concretely, to choose $\alpha_i$'s (and by extension $C_*$), \citet[Section 4.7.5]{boyd_convex_2004} recommend adjusting the relative ``weights'' $\alpha_i/\alpha_j$'s. In particular, a point with large curvature of the trade-off function (aka, the knee of the trade-off function) is a good point to reach a compromise between the various objectives; accordingly, finding the $C_*$ that achieves the knee point is recommended. We will demonstrate this process empirically in~\Cref{sec:guidance} under \textit{\nameref{par:choosing}}.

\subsection{Regulator's Incomplete Information and Estimation of Builder Penalty Scalars $\lambda_\fair$ and $\lambda_\priv$}
\label{sec:lambda-calib}
The penalty scalars $\lambda_\fair$ and $\lambda_\priv$ (hereon, $\lambda_*$) are builder parameters that regulators can have, at best, \textit{incomplete information} about~\citep{fudenberg_game_1991}. Regulators using \pp would need to estimate these parameters. In this section, we provide a systematic way to do so on a dataset they have access to.

From~\Cref{sec:game-formalization}, remember that $\lambda_*$ are scalars that map the monetary penalties to model builder's loss space.  Consider two models $\model_1$ and $\model_2$ that achieve the same fairness guarantee: $\Ls_\fair (\model_1) = \Ls_\fair (\model_2)$ (\textbf{A}). We require that the two models achieve the same overall builder loss: $\ell_\build(\model_1) \approx \ell_\build(\model_2)$ (\textbf{R}). Using~\Cref{eq:model_builder_loss}, we have:
$\lambda_\priv C_\priv = \frac{\err_\build(\model_2) - \err_\build(\model_1)}{\Ls_\priv(\model_1) - \Ls_\priv(\model_2)}$. Alternatively, if we parameterize our models with $\bs \in \mathcal{S}$ instead of $\model \in \mathcal{W}$, we have 
$\lambda_\priv C_\priv  = \frac{\err_\build(\bs_2) - \err_\build(\bs_1)}{\Ls_\priv(\bs_1) - \Ls_\priv(\bs_2)}$. If the regulator is calibrating their model, they cannot know the true value of $\lambda_\priv$ because $\err_\build$ is unknown to them; but $C_\priv$ is a parameter that they control. For the purposes of estimating $\lambda_\priv$, then, we can assume $C_\priv=1$. We have to resolve the resulting inaccuracy  of estimating $\lambda_\priv$ by choosing  appropriate $C_\priv$ later on (per~\Cref{sec:incentive-design}); but only if the specification is not met. We will demonstrate this in practice in~\Cref{sec:guidance} under~\textit{\nameref{par:enforcing}}.

To calibrate $\lambda_*$, we want to ensure our requirement (R) is met under condition (A), so we find models in $S_\gmm = \{\bs \mid \Ls_\fair (\bs) = \gmm \}$ where $S_\gmm$ are the set of models that achieve fairness gap $\gmm$ , clearly for two models $\bs_1$ and $\bs_2 \in S_\gmm$, our requirement is met. Thus, regulator's estimate $\hat{\lambda}_\priv$ of $\lambda_\priv$ is:
\begin{align}
  \label{eq:lambda_calibration}
  \hat{\lambda}_\priv = \frac{1}{C_\priv} \operatorname*{E}_{\gmm \in [0, 1]}\operatorname*{E}_{\bs_1, \bs_2\in S_\gmm} \left[ \frac{\err_\build(\bs_2) - \err_\build(\bs_1)}{\Ls_\priv(\bs_1) - \Ls_\priv(\bs_2)}\right];
\end{align}
with a similar expression for $\hat{\lambda}_\fair$ replacing $L_\priv$ with $L_\fair$.
\section{Simulating \specg with \pp}
\label{sec:results}
In this section, we simulate \specg with \pp. 
In our empirical evaluation,
we ask what we can learn from agents dynamics in \specg towards designing more effective ML regulations. Notably, we show single-agent-optimized regulations are ineffective in the multi-agent setting of ML regulation and that it benefits regulators to take initiative in specifying regulations.  Next, we show how regulators can choose their penalty scalars $C_*$ in practice (see~\Cref{sec:incentive-design}) and enforce compliance with their specifications despite their incomplete information regarding builder's penalty scalars $\lambda_*$ (see~\Cref{sec:lambda-calib}). Finally, we empirically validate that information equality is not necessary for \pp (see~\Cref{sec:assumption}). Our code for reproducing the experiments can be found at: \url{https://github.com/cleverhans-lab/ml_reg_games}.

\subsection{Experimental Setup}
\label{sec:exp-setup}
\paragraph{Algorithms}
We instantiate \pp with two algorithms: \fpate~\citep{impartiality} and \fDPbl~\citep{esipova_disparate_2023}. %
Both algorithms train fair and private classification models and adopt $(\varepsilon, \delta)$-DP
as their privacy notion.  \fpate adopts demographic parity which requires equalized prediction rates between different subgroups. Accordingly, we define $\boldsymbol{s}_\text{\fpate} = (\gmm, \eps),$ where $\gmm$ is the maximum tolerable demographic disparity between any two subgroups, and $\eps$ is the differential privacy budget. Furthermore, \fpate produces classifiers with a \textit{reject option}~\citep{ortner_learning_2016} which means that the classifier can reject answering queries (instead of producing inaccurate, or in this case, unfair) decisions. As a result, we also measure ``coverage'' as another utility metric besides accuracy which denotes the percentage of queries answered by the model at inference time. Higher coverage is better. \stephan{As rejection can also come at a cost (of invoking another model or deferring prediction to a human).}

\fDPbl~adopts disparate impact of DP~\cite{bagdasaryan_differential_2019} as its fairness notion which is the maximum accuracy gap between subgroups as a result of using DP. The aforementioned gap is then bounded by the tolerance threshold $\tau$. We define $\boldsymbol{s}_\text{\fDPbl} := (\tau, \eps)$ where $\eps$ is the privacy budget as before. We note that since the fairness violation is defined as the difference in disparate impact of privacy on accuracy between groups, it is a function of both $\tau$ and $\eps$. Consequently, when regulators update $\boldsymbol{s}$ in~\Cref{line:reg-update}~of~\Cref{alg:paretoplay}, they have to update both parameters.

 Our goal is to establish the general applicability of \specg and \pp. We do not compare the two aforementioned models given their differing fairness notions. 
 We run each experiment with 5 different initial specification $\bs_\reg$ %
  and aggregate the results. All results are plotted with 95\% confidence intervals.
 We defer further details to~\Cref{sec:pp-setup}. 

\paragraph{Datasets}
We adopt the experimental setup of \citet{impartiality} for \fpate and \citet{esipova_disparate_2023} for \fDPbl~as much as possible to provide comparable results. 
For \fpate, we perform gender classification on UTKFace~\citep{zhifei2017cvpr} and Fairface~\citep{karkkainenfairface} datasets where ``race'' is the sensitive attribute.
On CelebA~\citep{liu2015faceattributes} the classification task is ``whether the person is smiling'' and ``gender'' is the sensitive attribute~\citep{impartiality}. \citet{esipova_disparate_2023} evaluate~\fDPbl~on MNIST~\citep{deng2012mnist} for the typical 10-digit classification task. To force disparate behavior, they under-sample class 8 such that it constitutes about 1\% of the dataset on average, making it the underrepresented class. They report the the gap between disparate impact values on class 2 and class 8 as their (un)fairness measure which we adopt for our~\fDPbl~results.

\paragraph{\specg and \pp Settings}
All games are regulator-led unless specified otherwise. After each round of the game, we log the current strategies $\boldsymbol{s}$, the achieved privacy cost $\eps$, accuracy, and the achieved maximum demographic disparity for \fpate and disparate impact gap for \fDPbl. 
In~\Cref{sec:game-formalization}, we mentioned that in a repeated game, agent $i$ discounts their past losses over time by $c_i$. From an optimization point of view, $c_i$ appears as a decay factor (\Cref{line:decay} in~\Cref{alg:paretoplay}). We set $c_i = 1.5$ for all agents. %
We estimated builder's penalty scalars on UTKFace and FairFace to be $\lambda_\fair=0.3$ and $\lambda_\priv=0.01$ using the procedure detailed in \Cref{sec:lambda-calib}.

\subsection{Results: Agent Dynamics and Policy Guidance for ML Regulators}
\label{sec:guidance}
\begin{figure}[h]
    \centering
     \begin{subfigure}[b]{0.24\textwidth}
         \centering
         \includegraphics[width=\textwidth]{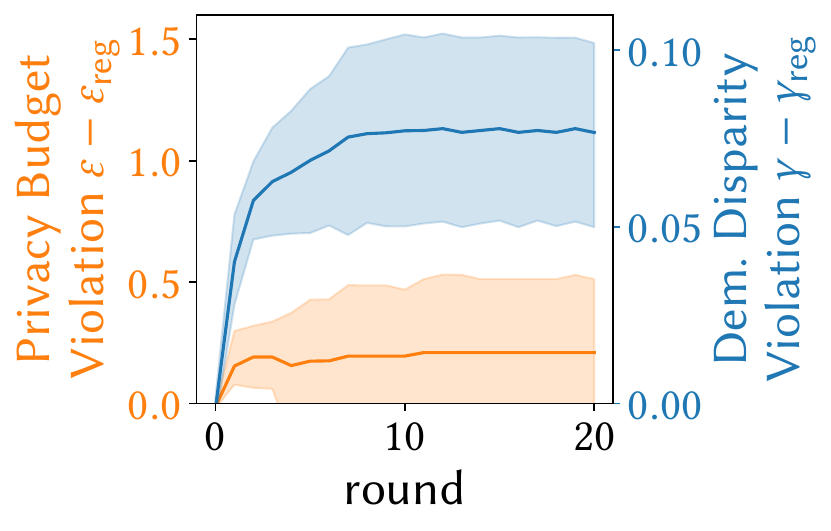}
         \caption{\fpate on UTKFace}
         \label{fig:RQ1_utkface}
     \end{subfigure}
     \begin{subfigure}[b]{0.24\textwidth}
         \centering
         \includegraphics[width=\textwidth]{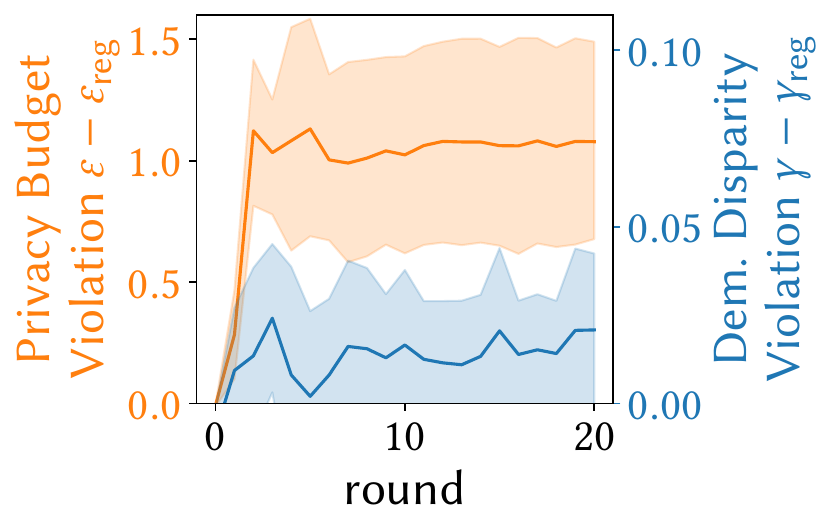}
         \caption{\fpate on CelebA}
         \label{fig:RQ1_celeba}
     \end{subfigure}
     \begin{subfigure}[b]{0.24\textwidth}
         \centering
         \includegraphics[width=\textwidth]{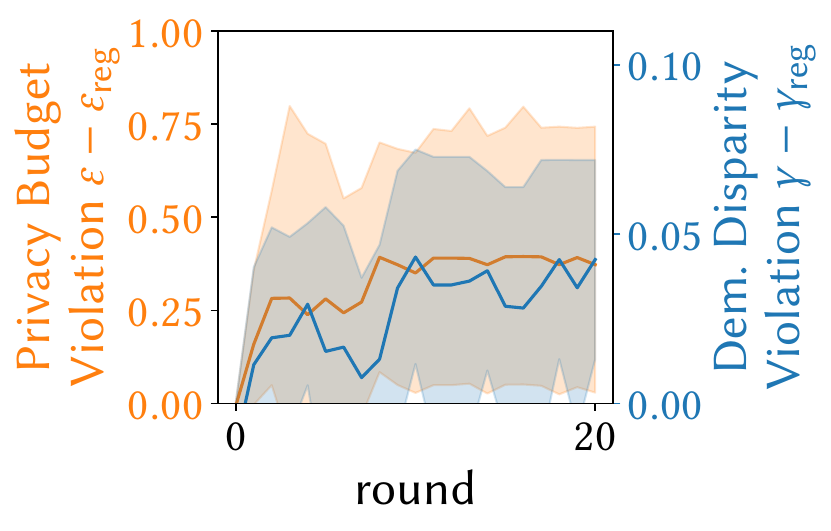}
         \caption{\fpate on FairFace}
         \label{fig:RQ1_fairface}
     \end{subfigure}
     \begin{subfigure}[b]{0.24\textwidth}
         \centering
         \includegraphics[width=\textwidth]{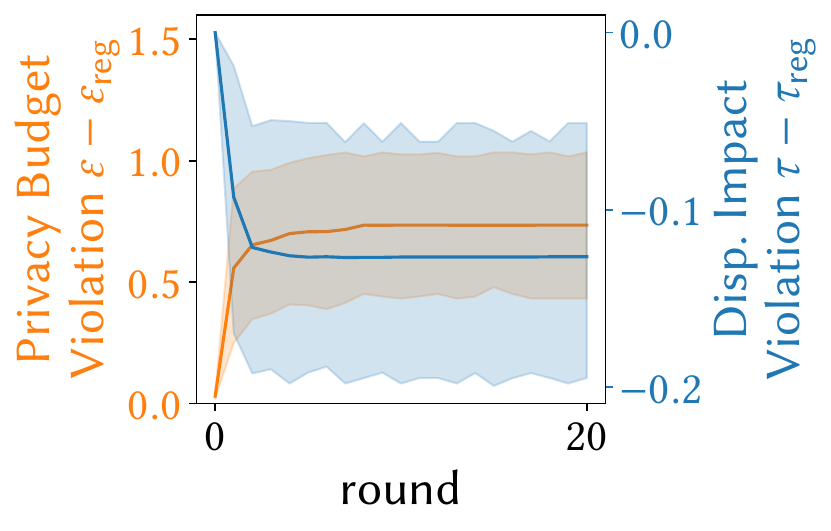}
         \caption{DPSGD-G.-A. on MNIST}
         \label{fig:RQ1_mnist}
     \end{subfigure}
        \caption{\textbf{Single-agent-optimized regulations are ineffective.} In regulator-led games, regulators specify $\bs_\reg = (\gamma_\reg, \eps_\reg)$ expecting builders produce models that satisfy $\bs - \bs_\reg \approx 0$. We show privacy ($\varepsilon$) and fairness ($\gamma$) violations from regulators' specification in orange (left) and blue (right) axis, respectively. In the absence of penalties, utility-seeking builders (regardless of algorithm or dataset) eventually violate the specifications. The confidence region is 95\% using 5 different values for regulator's specifications $\bs_\reg$. 
        Results on \fDPbl~(d) show improvement in fairness beyond regulator's specification but this comes at the cost of privacy.}
        \label{fig:RQ1}
\end{figure}

\paragraph{Single-agent-optimized regulations are ineffective}
As a baseline technique for producing regulations, consider the scenario where regulators optimize regulation without consideration for the builder's response. In other words, regulators specify $\bs_\reg$ expecting that builders follow their specification and thus announce no penalties.
In~\Cref{fig:RQ1}, we show that in this scenario \specg with both \fpate and \fDPbl~converge to equilibria that violate regulators' specified guarantees.
In all four games, the regulators act first and choose strategies $\bs$ that produce models that initially meet their privacy and fairness (disparity) guarantees. However, in the absence of penalties, the model builders modify the model parameters to improve their own utility.

Each game converges to a point that either violates the fairness or the privacy specification. Notably, \fDPbl~(\Cref{fig:RQ1}-d) shows improvement in fairness beyond the regulator's specification. This is a consequence of the fairness notion (disparate impact of DP) which aligns fairness regulator and model builder objectives. 
However, this comes at the cost of privacy specification which is unacceptable.

In the rest of the experiments, we focus on the dynamics of \specg and show how to estimate its various parameters. In the interest of a more focused presentation,
we will adopt \fpate as the underlying fair and private learning algorithm, and use ``disparity'' instead of ``demographic disparity,'' for conciseness.

\paragraph{\specg leader has a first-mover advantage} 

\begin{figure}[htp]
     \centering
     \includegraphics[width=0.48\textwidth]{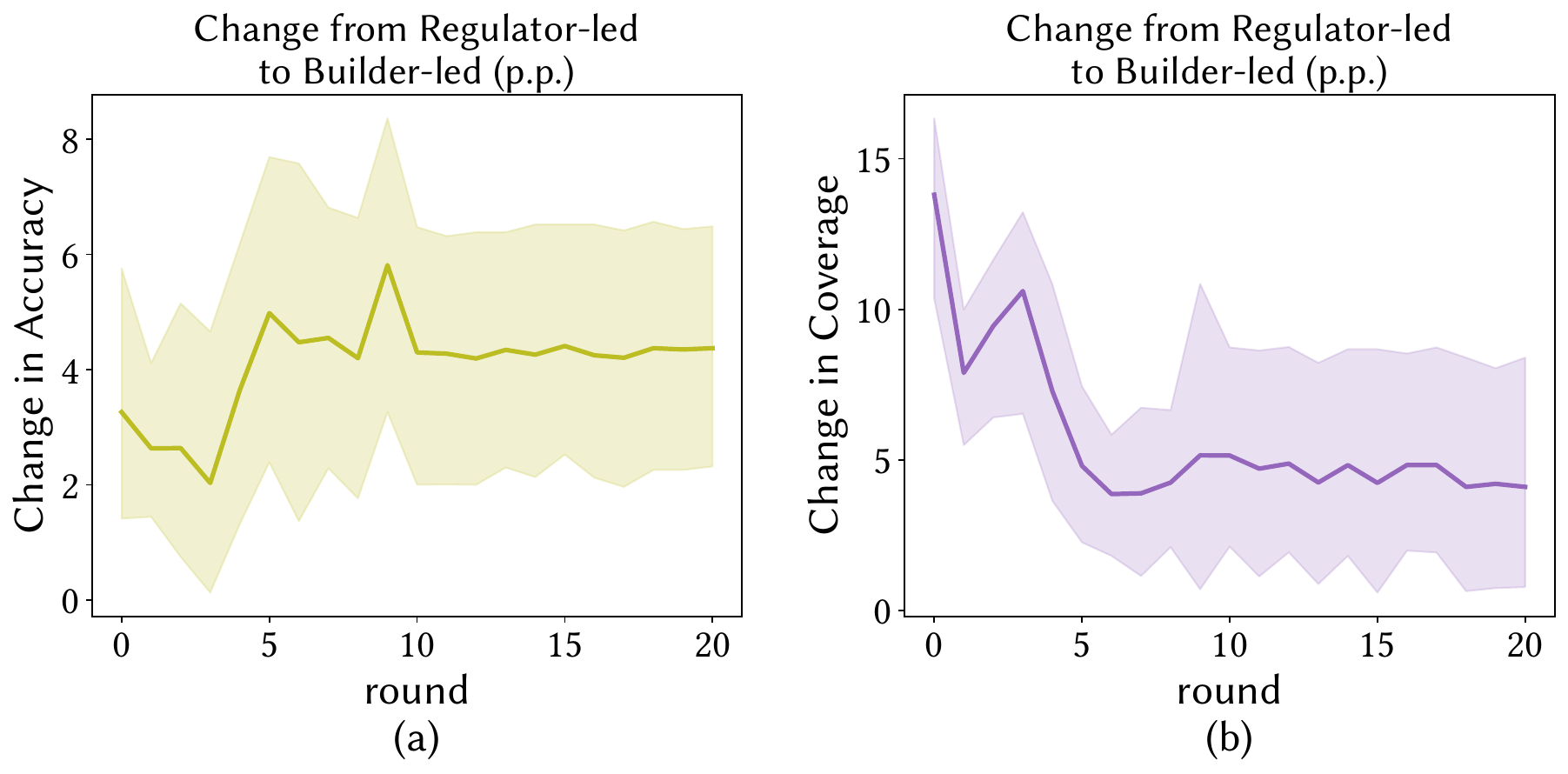}
     \hfill
     \includegraphics[width=0.49\textwidth]{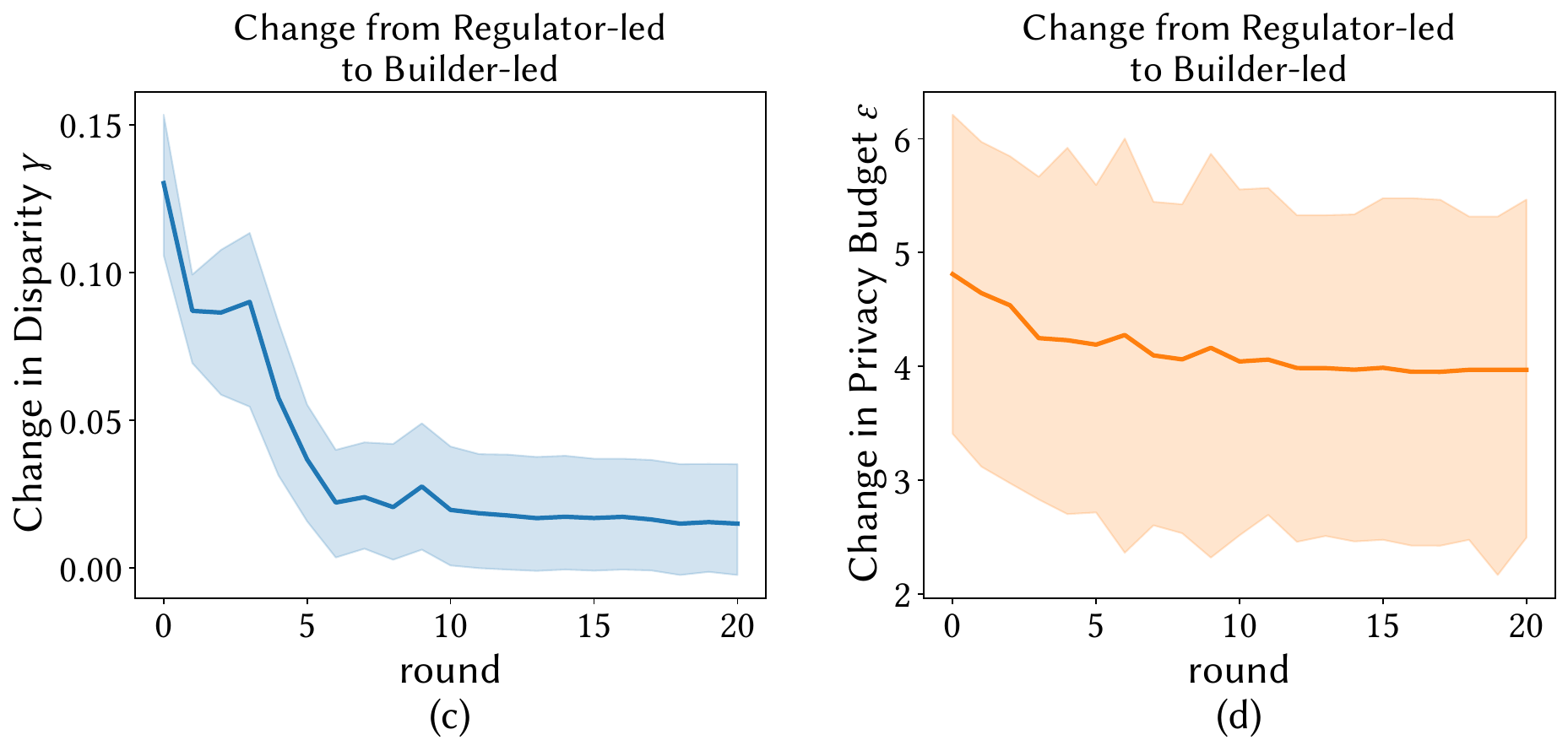}
        \caption{\textbf{First-mover has an advantage in \specg.} 
        We compare a builder-led game to a regulator-led one and show the differences in objective values. When the builder leads,
        it produces models that are on-average 5 percentage points more accurate (a) and answer 5 percentage points more queries (b) compared to when the regulator leads; however, this comes at the cost of 0.02 increase in disparities (c) and a privacy budget increase of 4 (d).} %
        \label{fig:RQ2_1}
\end{figure}

Recall that in each game, the first-mover chooses the point on the Pareto surface that minimizes their loss. All other parameters in both games, including regulators' fairness and privacy constraints, remain the same throughout the game run. In~\Cref{fig:RQ2_1}, we show the difference in achieved objective values changing from a regulator-led game to an builder-led one. When the builder leads,
it produces models that are on-average 5 percentage points more accurate (a) and answer 5 percentage points more queries (b) compared to when the regulator leads; however, this comes at the cost of a minor 0.02 increase in disparities (c) and but a large privacy budget increase of 4 (d). Therefore, regulators should take initiative in making their specifications.

\begin{figure}
\includegraphics[width=\linewidth]{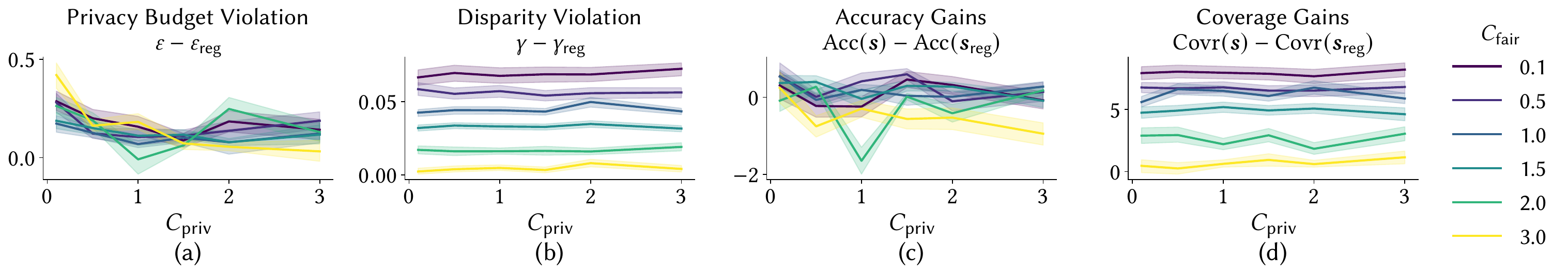}
\caption{
\textbf{Choosing $C_\fair$ and $C_\priv$}. We run regulator-led games with different $C_\fair$ and $C_\priv$ combinations. We specification violations (a-b) and utility gains (c-d) over 20 rounds of game as a function of $C_\priv$ and represent $C_\fair$ with different hues. (a) $C_\priv$ = 1.5 is a good choice since it is the knee-point for most $C_\fair.$ (b) $C_\fair= 3.0$ reduces fairness violation to 0 without sacrificing builder's utility too much.
}
\label{fig:C_combinations}
\end{figure}%

\paragraph{Choosing $C_\fair$ and $C_\priv$}
\label{par:choosing}
In~\Cref{fig:C_combinations}, we show violations from regulator specification as a function of its penalty scalars $C_\priv$ (x-axis) and $C_\fair$
(hue) using \fpate on UTKFace dataset. Confidence intervals are calculated over 20 rounds of \specg starting from 5 different specifications. We use pre-calibrated $\lambda_\fair=0.3$ and $\lambda_\priv=0.01$ per~\Cref{sec:lambda-calib}. %
Our goal is to reduce violations to near zero without unnecessarily harming builder utility (coverage and accuracy). Looking at privacy violations (\Cref{fig:C_combinations}-a), $C_\priv=1.5$
is a sensible choice as it is the knee-point for most $C_\fair$ choices, coupled with $C_\fair=3$ (\Cref{fig:C_combinations}-b) it ensures zero disparity violations and no hit to accuracy (\Cref{fig:C_combinations}-c). Its only trade-off is with coverage (\Cref{fig:C_combinations}-d), where choosing smaller $C_\fair$ can allow for up to 8\% higher coverage (more queries answered by the builder) but which comes clearly at the expense of fairness (\Cref{fig:C_combinations}-b). We note that excessively large values for $C_\priv$ (especially with small $C_\fair$ in \Cref{fig:C_combinations}-a) can push us to undesirable equilibria with on average $0.25$ higher privacy budgets. This non-linear behavior demonstrates that effective values for $C_*$ are not only a function of $\lambda_*$ (decided by the builder), but are also influenced by the inherent trade-offs between the various objectives on the ML task.

\paragraph{Enforcing equilibria despite incomplete information}
\label{par:enforcing}

\begin{figure}
     \centering
         \centering
         \includegraphics[width=\textwidth]{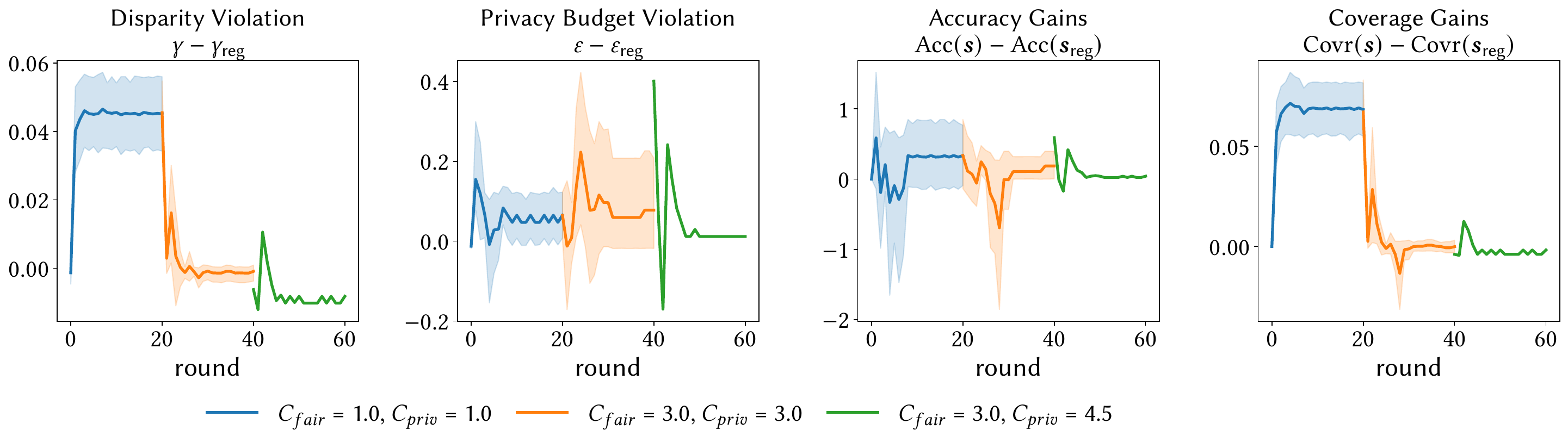}
        \caption{\textbf{Regulators can enforce desired equilibria despite incomplete information.} 
        We show a scenario where initial penalties were ineffective in enforcing compliance with the specification (blue) due to incomplete information about builder's loss. Regulators re-calculate their penalty scalars $C_\fair, C_\priv$ to progressively enforce stronger penalties in two subsequent phases of the game (orange and green) with the goal of reducing the number of violations. Games are regulator-led and there are 5 different initial specifications.
        }
        \label{fig:RQ3}
\end{figure}

Exogenous factors aside, given the uncertainty regarding the builder's dataset and its parameters $\lambda_*$, it is possible that penalties issued are not enough to avoid specification violations. If the game has converged to an undesirable equilibrium, regulators can change their penalty scalars $C_*$ to enforce their specification accordingly. We demonstrate this in
\Cref{fig:RQ3}. The game has multiple phases in each of which we run \specg until convergence. We simulate the aforementioned uncertainty by assuming no priors on $C_*$, and choosing $C_\fair=1.0$ and $C_\priv=1.0$ in the first phase. As before, we simulate the outcome for 5 %
different initial specifications $\bs_\reg$ using which we draw the 95\% confidence intervals. In the first phase (blue), we observe that a large portion of the games violate disparity specifications by 6\% for a similar improvement in coverage. The constraint violations are due to inappropriate penalties. %
In the second phase (orange) we recalculate $C_\fair=C_\priv = 3.0$ which manages to reduce fairness violations to 0. In this phase we get more consistent adherence with the fairness specification, but larger violations of privacy. Finally, we are left with one violation of privacy specification, increasing $C_\priv$ to 4.5 allows us to enforce that specification as well (green).

\begin{figure}[htp]
     \centering
        \includegraphics[width=\textwidth]{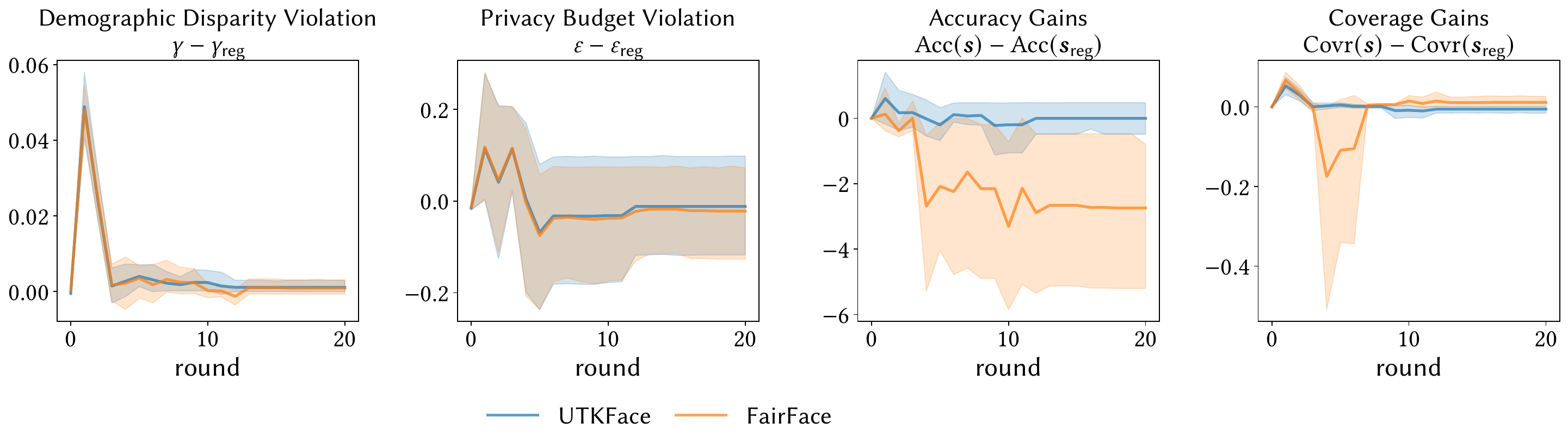}
        \caption{\textbf{Agents can have separate datasets in \pp.} We simulate a regulator-led game where regulators have access to FairFace and model builder has access to UTKFace. The resulting builder's model has on average 2\% higher accuracy compared to the regulator's. Despite these differences, \specg converges and follows a similar trajectory for both agents in terms of privacy and fairness violations---ensuring that regulator specifications are generally satisfied. The 95\% confidence intervals are over 5 different initial specifications.}
        \label{fig:two_datasets}
\end{figure}

\paragraph{Information equality is not necessary for \pp}
\label{par:assumption}
We empirically validate the assumption of a shared \pf between regulators and model builder, despite them having access to different datasets (see~\Cref{sec:assumption}). In~\Cref{fig:two_datasets}, regulators have access to FairFace, whereas the model builder has access to UTKFace. The agents use their respective dataset to form their loss functions. Each trains and calibrates their own model on their own datasets, and therefore, calculate their own contributions to the shared \pf. The results of the game is shown in \Cref{fig:two_datasets}. The builder's model has on average 2\% higher accuracy compared to the regulator's. However, \specg converges and follows a very similar trajectory for both agents in terms of privacy and fairness violations---ensuring that regulator specifications are generally satisfied. 

\section{Related Work}
\label{sec:related}

It has been shown that private training of ML models negatively influences utility~\citep{tramer2020differentially} and fairness~\citep{vinith, farrand2020neither}.
A general line of work on integrating those objectives by adapting the training procedure~\citep{xu2019achieving,mozannar2020fair,franco2021toward,tran2021lagrangian}, or identifying favorable trade-offs between subsets of these objectives~\citep{avent2019automatic} has emerged over the past years.
Note that, in contrast to our work, all prior frameworks to unify different objectives or characterize trade-offs do not consider the inherent multi-agent nature of the problem.
Yet, we can leverage their methods to instantiate our regulation games.
In this work we build on two recent frameworks, namely FairPATE~\citep{impartiality} and DPSGD-Global(-Adapt)~\citep{esipova_disparate_2023}.
The fact that our regulation games can be instantiated with two frameworks that differ so significantly in their approach to integrate privacy and fairness highlights the universality of our work.
In work closest to ours,  \citet{jagielski2019differentially} study the trade-offs between privacy, fairness, and accuracy within a game theoretic framework through a two-player zero-sum game. Our focus is on formulating the regulation game, with the purpose of designing proper incentive. We are interested in the more general (and more realistic) case of having multiple agents (such as two regulatory bodies) interacting with the model owner which does not admit a two-player zero sum game solution.

\section{Limitations and Discussion}
With the increasing importance of machine learning in sensitive domains, it is crucial to ensure that the machine learning models are trustworthy. 
However, previous research has primarily focused on addressing a single trust objective at the time or, when considering multiple objectives, assumed the existence of a central entity responsible for implementing all objectives. 
We highlight the limitations of this assumption for realistic scenarios with multiple agents and introduce an approach for optimization over multiple agents with multiple objectives to overcome this limitation.

Our approach recognizes the diverse nature of agents involved in deploying and auditing machine learning models. This allows us to make suggestions for guarantee levels that are more likely to be realizable in practice; given that the gains and benefits of different parties have been taken into account. We, however, acknowledge that agents may in fact have a more diverse set of requirements and objectives; and that as a result our models may not be sophisticated-enough to incorporate all such factors. Additionally, we made several assumptions regarding the economic model under which we operate as well as common knowledge of the \pf between various objectives. While these assumptions follow established principles in economics (expected utility hypothesis for the former) and  in machine learning (the existence of a data-generating distribution for the latter), both are contested in their respective literature.  

Finally, we acknowledge that providing ``metrics'' for human and society values such as fairness and privacy is imperfect at best and fraught with philosophical and ethical issues. Nevertheless, the metrics we used in our study are commonplace in trustworthy ML circles and the search for better, more inclusive, metrics is underway. Our research, therefore, aims to provide systematic guidance on best practices in regulating trustworthy ML practices, and can be adopted for future development in these areas.

From our empirical results, we observe that different ML tasks exhibit different Pareto frontiers (see~\Cref{app:pareto-frontiers}). As such, an \specg played for one task cannot necessarily provide regulation recommendation for other tasks. It remains to be seen how much such recommendations can transfer between tasks even within the same domain (for instance, vision). For instance, recommendation made on the basis of age classification may be ineffective (or too restrictive) for gender estimation.

Finally, we centered our consideration around calculating fines proportional to the privacy and fairness violations of 
chosen guarantee levels ($\gamma, \varepsilon$); as well as ensuring they are effective in changing model builder behavior. The converse problem is also important: assuming a bound $C$ on the penalty, what are the maximal $\gamma, \varepsilon$ guarantees that we can expect to be able to enforce?

\section*{Acknowledgements}
We would like to acknowledge our sponsors, who support our research with financial and in-kind contributions: Amazon, Apple, CIFAR through the Canada CIFAR AI Chair, DARPA through the GARD project, Intel, Meta, NSERC through a Discovery Grant, the Ontario Early Researcher Award, and the Sloan Foundation. Resources used in preparing this research were provided,
in part, by the Province of Ontario, the Government of Canada through CIFAR, and companies sponsoring the Vector Institute.
  
\printbibliography

\clearpage

\appendix
\label{app}
\section{Notation}
\begin{table}[H]
    \centering
    \begin{tabular}{cl}
        \toprule
        \textbf{Notation} & \textbf{Explanation} \\
        \midrule
        $\model$ & Model \\ 
        $\Gamma$ & Fairness metric \\
        $\ell_\fair(\gmm; \model)$ & Fairness regulator loss of model $\model$ with constraint $\gmm$ \\
        $\gmm$ & Fairness gap \\
        $\hgmm(\model)$ & Measured fairness gap on model $\model$ \\
        $\ell_\priv(\model)$ & Privacy regulator loss of model $\model$ \\
        $\eps$ & Privacy cost \\
        $\heps(\model)$ & Estimated privacy cost on model $\model$ \\
        $\err_\build(\model; D)$ & Model builder (utility) loss on dataset $D$\\
        $\bs$ & strategy (specification)\\
        $C_{*}$ & regulator penalty scalar \\
        $\eta$ & step size used by model builder for updates \\
        $c_{*}$ & step size discounting factor \\
        \bottomrule
    \end{tabular}
    \caption{Table of Notation}
    \label{tab:notation}
\end{table}
\section{Background on Game Theory}
\label{app:game_theory}

\subsection{Best Response}
\label{app:best-response}
In a multi-agent setup where every agent cares about only one objective (its own),
we have a \textit{bi-level optimization} problem. For instance the model builder would be solving:
\begin{equation}
\label{eq:opt2}
\begin{array}{cc}
\min_{\theta_{\text {acc}}} & \ell_{\text {acc}}\left(\theta_{\text {acc}}, \theta_{\text {priv}}, \theta_{\text {fair}}\right) \\ 
\text { subject to } & \theta_\text{priv} = \arg\min_{\theta_\text{priv}} \ell_{\text{priv}}\left(\theta_{\text {acc}}, \theta_{\text {priv}}, \theta_{\text {fair}}\right) \\
& \theta_\text{fair} = \arg\min_{\theta_\text{fair}} \ell_{\text {fair}}\left(\theta_{\text {acc}}, \theta_{\text {priv}}, \theta_{\text{fair}}\right)
\end{array}
\end{equation}

Consider $\bar{\theta} = \begin{bmatrix} \theta_\text{acc} &\theta_\text{priv} &\theta_\text{acc}\end{bmatrix}^\top$, then we can write the objectives of all agents as $$\bar{\ell} (\theta) := 
    \begin{bmatrix}
    \ell_\text{acc}\left(\bar{\theta}\right) &\ell_\text{priv}\left(\bar{\theta}\right) &\ell_\text{fair}\left(\bar{\theta}\right)
\end{bmatrix}^\top.$$ The map $\bar{\ell}: \Theta_{\text {acc}} \times \Theta_{\text {priv}} \times \Theta_{\text {fair}} \mapsto \mathbb{R}_{\geq 0}^3$ is vector-valued.  Let $\bar{\theta^*}$ be the solution to the bi-level optimization of \cref{eq:opt2}: 

\begin{equation}
\label{eq:game-opt}
\bar{\theta^*} = \arg\min_{\bar{\theta}} \begin{bmatrix}
     \ell_\text{acc}\left(\bar{\theta}\right)
    &\ell_\text{priv}\left(\bar{\theta}\right)
    &\ell_\text{fair}\left(\bar{\theta}\right)
\end{bmatrix}
\end{equation}

This formulation allows us to study the interaction of agents whose objectives are defined through a value function (known as the \textit{payoff}). 
Note that a agent's payoff is not only a function of its own actions, but also those of its peers. This creates an opportunity for the agent to strategize and choose its best possible action given others' actions, where ``best'' is interpreted as the optimizer of its payoff. These actions form the \textit{best responses} (or BRs) to peers' actions. Therefore, BRs are set-valued mappings from the set of agents' actions onto itself whose fixed points are known as \textit{Nash equilibria}. 

More formally, let $BR: \Theta_{\text {acc}} \times \Theta_{\text {priv}} \times \Theta_{\text {fair}} \mapsto \Theta_{\text {acc}} \times \Theta_{\text {priv}} \times \Theta_{\text {fair}}$ be the argmin function of $\bar{\ell}$. This operator calculates the \textit{best response} (BR) of every agent given the choice of parameters $\theta$. $\bar{\theta^*}$ is \textit{fixed-point} of this map:
    $BR(\bar{\theta^*}) = \bar{\theta^*}$. $\bar{\theta^*}$ is a Nash Equilibrium and \cref{eq:game-opt} describes a game.

\subsection{Stackelberg Competitions}
\label{app:game_stackelberg}
Stackelberg competitions model sequential interaction among strategic agents with distinct objectives~\cite{fudenberg_game_1991}.
They involve a leader and a follower. The leader is interested in identifying the best action (BR) assuming rational behavior of the follower.
The combination of the leader's action and the follower’s rational best reaction leads to a strong Stackelberg equilibrium (SSE)~\cite{birmpas2020optimally}.
This improves over work relying on zero-sum game formulation~\cite{yao1977probabilistic} where the follower's objective is assumed to be opposed to the leader's objective.
An important example for the application of Stackelberg competition in trustworthy ML strategic classification.
Therein, strategic individuals can, after observing the model output, adapt their data features to obtain better classification performance.
Such changes in the data can cause distribution shifts that degrade the model's performance and trustworthiness on the new data, and thereby requires the model builders adapt their models. In our model governance game framework, the two regulators act as \emph{leaders} while the model builder acts as the \emph{follower}.
By following the Stackelberg competition, the model builder aims at obtaining the best-performing ML model given the requirements specified by the regulators.

\section{\pp Setups}
\label{sec:pp-setup}
\subsection{\pp on FairPATE}
\label{sec:pp-fairpate}
In \fpate, we train teacher ensemble models on the training set. These teachers  vote to label the unlabeled public data. We then train student models on the now labeled public data. At inference time, the student model does not answer all the queries in the test set. It refrains from answering a query when answering it would violate the fairness constraint. Coverage measures the percentage of queries that the student does answer. 

We denote the student model for classification by $\model$, the features as $(\mathbf{x}, z) \in \mathcal{X} \times \mathcal{Z}$ where $\mathcal{X}$ is the domain of non-sensitive attributes, $\mathcal{Z}$ is the domain of the sensitive attribute (categorical variable).
The categorical class-label is denoted by $y \in [1, \dots, K]$. We indicates the strategy vector space as $\bs = (\gmm, \eps)$ where $\gmm$ is the maximum tolerable fairness violation and $\eps$ is the privacy budget.

The loss functions of all agents depend on both $\gmm$ and $\eps$. A gradient descent update of $\gmm$ and $\eps$ is:
\begin{align} 
\gmm^t = \gmm^{t-1} - \eta_\text{fair}\frac{\partial{\Ls}}{\partial{\gmm}}, \;
\eps^t = \eps^{t-1} - \eta_\text{priv}\frac{\partial{\Ls}}{\partial{\eps}}
\end{align}

The model builder cares about both student model accuracy and coverage. It would want to provide accurate classification and answer most queries. Its loss function uses a weighted average of the two:
\begin{equation}
\ell_\text{b}(\gmm, \eps) = -\left(\lambda_{b}\acc(\gmm,\eps) + (1-\lambda_{b})\cov(\gmm,\eps)\right)    
\end{equation}
where $\lambda_{b}$ is a hyperparameter set by the model builder that controls how much it values accuracy and coverage. The accuracy and coverage are multiplied with -1 to form the loss because we want to maximize them. Both accuracy and coverage values used are between 0 and 1.  At each turn, the model builder decides its response by calculating $\frac{\partial{\ell_b}}{\partial{\gmm}}$ and $\frac{\partial{\ell_b}}{\partial{\eps}}$ at the current $\gmm$ and $\eps$. In our experiments, we use $\lambda_b = 0.7$.

The loss function of the fairness and privacy regulators are $\ell_\text{fair}(\gmm,\eps) = \gmm_\text{ach}(\gmm,\eps) ~\text{and}~ \ell_\text{priv}(\gmm,\eps) = \eps_\text{ach}(\gmm,\eps)$ respectively.

\subsection{\pp on DPSGD-Global-Adapt}
In \fDPbl, we train instances of private models with $\eps$ and thresholds $\tau$ on MNIST. We then also train an non-private model on MNIST with the same training set to compute the fairness gap using \ref{eq:accuracy_gap}.

We indicates the strategy vector space as $\bs = (\tau, \eps)$ where $\tau$ is the maximum tolerable fairness violation and $\eps$ is the privacy budget. We make a slight modification to the original algorithm to accept privacy budget $\eps$ instead of a noise multiplier $\sigma$ in order to make the presentation consistent with the rest of our results.

Similarly, the loss functions of all agents depend on both $\tau$ and $\eps$. A gradient descent update of $\gmm$ and $\eps$ is:
\begin{align} 
\tau^t = \tau^{t-1} - \eta_\text{fair}\frac{\partial{\Ls}}{\partial{\tau}}, \;
\eps^t = \eps^{t-1} - \eta_\text{priv}\frac{\partial{\Ls}}{\partial{\eps}}
\end{align}

As before, we assume the model builder primary concern is higher model accuracy. Its loss function is:
\begin{equation}
\ell_\text{b}(\tau, \eps) = -\acc(\gmm,\eps)     
\end{equation}
\section{Fairness}
\label{sec:fairness}
We provide more details on the fairness notions used in our empirical study in~\Cref{sec:results}.

\subsection{Demographic Parity Fairness}
\citet{impartiality} adopt the the fairness metric of \emph{multi-class demographic parity} which requires that ML models produce similar success rates (\ie, rate of predicting a desirable outcome, such as getting a loan) for all subpopulations~\citep{calders2010three}. 

In practice, they estimate multi-class demographic disparity for class $k$ and subgroup $z$ with:
$
    \widehat{\Gamma}(z, k) := 
    \frac{|\{\hat{Y}=k, Z = z \}|}{|\{Z = z\}|} - \frac{|\{\hat{Y}=k, Z \neq z\}|}{|\{Z \neq z\}|},
$
where $\hat{Y} = \model(\mathbf{x}, z)$. They define demographic \textit{parity} when the worst-case demographic disparity between members and non-members for any subgroup, and for any class is bounded by $\gamma$: %
\begin{definition}[$\gamma$-DemParity] For predictions $Y$ with corresponding sensitive attributes $Z$ to satisfy $\gamma$-bounded demographic parity ($\gamma$-DemParity), it must be that for all $z$ in $\mathcal{Z}$ and for all $k$ in $\mathcal{K}$, the demographic disparity is at most $\gamma$: $\Gamma(z, k) \leq \gamma$.
\end{definition}

\subsection{Disparate Impact of Differential Privacy}
\citet{esipova_disparate_2023} study the disparate impact of privacy on learning across different groups. 
In particular, they adopt a \textit{accuracy parity} notion of fairness~\citep{bagdasaryan_differential_2019}. 
A fair model, in their view, minimizes the gap between disparate impact values across groups. The disparate impact of privacy is defines as the following:
\begin{equation}
\pi\left(\model, D_k\right)=\operatorname{acc}\left(\model^* ; D_k\right)-\mathbb{E}_{\tilde{\model}}\left[\operatorname{acc}\left(\tilde{\model} ; D_k\right)\right],
\label{eq:accuracy_disparate}
\end{equation}
where $\acc\left(\model^* ; D_k\right)$ is accuracy of the optimal accuracy $\model^*$ on dataset $D_k$ belonging to the $k$\textsuperscript{th} subpopulation, and $\tilde{\model}$ is the privatized model. As with the output of any differentially private mechanism, the accuracy of the privatized model is measured in expectation. 

In our experiments, we follow their definition and measure the gap between class 2 and class 8 in~\Cref{eq:accuracy_gap}. That is, the fairness measure for the regulator in the regulator objective when using~\fDPbl~is:
\begin{equation}
    \Gamma_\text{\fDPbl}(\model; D) = \pi_{2, 8} = |\pi_2 - \pi_8|
\label{eq:accuracy_gap}
\end{equation}

\section{Additional Experimental Setup}
\label{sec:setup}
In all games, we set step size discount factor to $c = 0.67$. For \fpate, we use step sizes $\eta_\text{fair} = 0.1$ and $\eta_\text{priv} = 10$. We set model builder's loss function weightings to $\lambda_\text{fair} = 0.3$, $\lambda_\text{priv} = 0.01$, and $\lambda_\text{b} = 0.7$. For \fDPbl, we use step sizes $\eta_\text{fair} = 0.1$ and $\eta_\text{priv} = 5$. All the other game hyperparameters for each game shown in \Cref{sec:results} are shown in \Cref{tab:additional_setup}.  

The model architecture and data we use in the experiments follow what is described in the original works for \fpate~\cite{impartiality} and \fDPbl~\cite{esipova_disparate_2023}. 
The datasets used for \fpate and their information are shown in \Cref{tab:datasets_fairPATE}. For all datasets in \fpate for the calibration step, we train the student model with Adam optimizer and binary cross entropy loss. We train for 30 epochs on UTKFace, 15 on CelebA, 25 on FairFace, and 60 on MNIST.

During the games, we put box constraints on the parameters $\bs = (\gmm, \eps)$ so that they would not be out of range and produce undefined outputs. We use $\gmm \in [0.01, 1]$ and $\eps \in [1, 10]$. 

\begin{table}[htp]
\centering
\begin{tabular}{l|ccccc}
    \toprule
    \textbf{Figure} & \textbf{Dataset} & \textbf{Algorithm} & \textbf{$C_{\fair}$} & \textbf{$C_{\priv}$} \\
    \midrule
    \Cref{fig:RQ1_utkface} & UTKFace & \fpate & 0 & 0 \\
    \Cref{fig:RQ1_celeba} & CelebA & \fpate & 0 & 0 \\
    \Cref{fig:RQ1_fairface} & FairFace & \fpate & 0 & 0 \\
    \Cref{fig:RQ1_utkface} & MNIST & DPSGD-Global-Adapt & 0 & 0 \\
    \Cref{fig:RQ2_1} & UTKFace & \fpate & 3 & 3 \\
    \Cref{fig:RQ2_1} & UTKFace & \fpate & 3 & 3 \\
    \Cref{fig:C_combinations} & UTKFace & \fpate & 0.1-3 & 0.1-3 \\
    \Cref{fig:RQ3} Round 1 & UTKFace & \fpate & 1 & 1 \\
    \Cref{fig:RQ3} Round 2 & UTKFace & \fpate & 3 & 3 \\
    \Cref{fig:RQ3} Round 3 & UTKFace & \fpate & 4.5 & 3 \\
    \Cref{fig:two_datasets} & UTKFace, FairFace & \fpate & 3 & 3 \\
    \bottomrule
\end{tabular}
\caption{\textbf{ParetoPlay hyperparameter settings used in the experiments.}}
\label{tab:additional_setup}
\end{table}

\begin{table}[htp]
\centering
\resizebox{\textwidth}{!}{
\begin{tabular}{lccccccccccc}
\toprule
Dataset & Prediction Task & C & Sens. Attr. & SG & Total &  U & Model & Number of Teachers & $T$ & $\sigma_1$ & $\sigma_2$  \\
\midrule
    CelebA & Smiling & 2 & Gender & 2 & \numprint{202599} & \numprint{9000} & Convolutional Network (\Cref{tab:CNN-CelebA})& 150 & 130 & 110 & 10\\
    FairFace & Gender & 2& Race& 7& \numprint{97698} & \numprint{5000} & Pretrained ResNet50 & 50 & 30 & 30 & 10\\
    UTKFace & Gender & 2 & Race & 5 & \numprint{23705} & \numprint{1500} & Pretrained ResNet50 & 100 & 50 & 40 & 15\\
\bottomrule
\end{tabular}
}
\caption{Datasets used for \fpate. Abbreviations: \textbf{C}: number of classes in the main task; \textbf{SG}: number of sensitive groups; \textbf{U}: number of unlabeled samples for the student training . Summary of parameters used in training and querying the teacher models for each dataset.
The pre-trained models are all pre-trained on ImageNet. We use the most recent versions from PyTorch. 
}
\label{tab:datasets_fairPATE}
\end{table}

\begin{table}[htp]
    \centering
    \begin{tabular}{lc}
    \toprule
    Layer & Description \\
    \midrule
    Conv2D & (3, 64, 3, 1)  \\
    Max Pooling & (2, 2) \\
    ReLUS & \\
    Conv2D & (64, 128, 3, 1)  \\
    Max Pooling & (2, 2) \\
    ReLUS & \\
    Conv2D & (128, 256, 3, 1)  \\
    Max Pooling & (2, 2) \\
    ReLUS & \\
    Conv2D & (256, 512, 3, 1)  \\
    Max Pooling & (2, 2) \\
    ReLUS & \\
    Fully Connected 1 & (14 * 14 * 512, 1024) \\
    Fully Connected 2 & (1024, 256) \\
    Fully Connected 2 & (256, 2) \\
    \bottomrule
    \end{tabular}
    \caption{Convolutional network architecture used in CelebA experiments.}
    \label{tab:CNN-CelebA}
\end{table}
\section{Additional Empirical Results}

\begin{figure}[htp]
    \centering
    \includegraphics[width=0.6\textwidth]{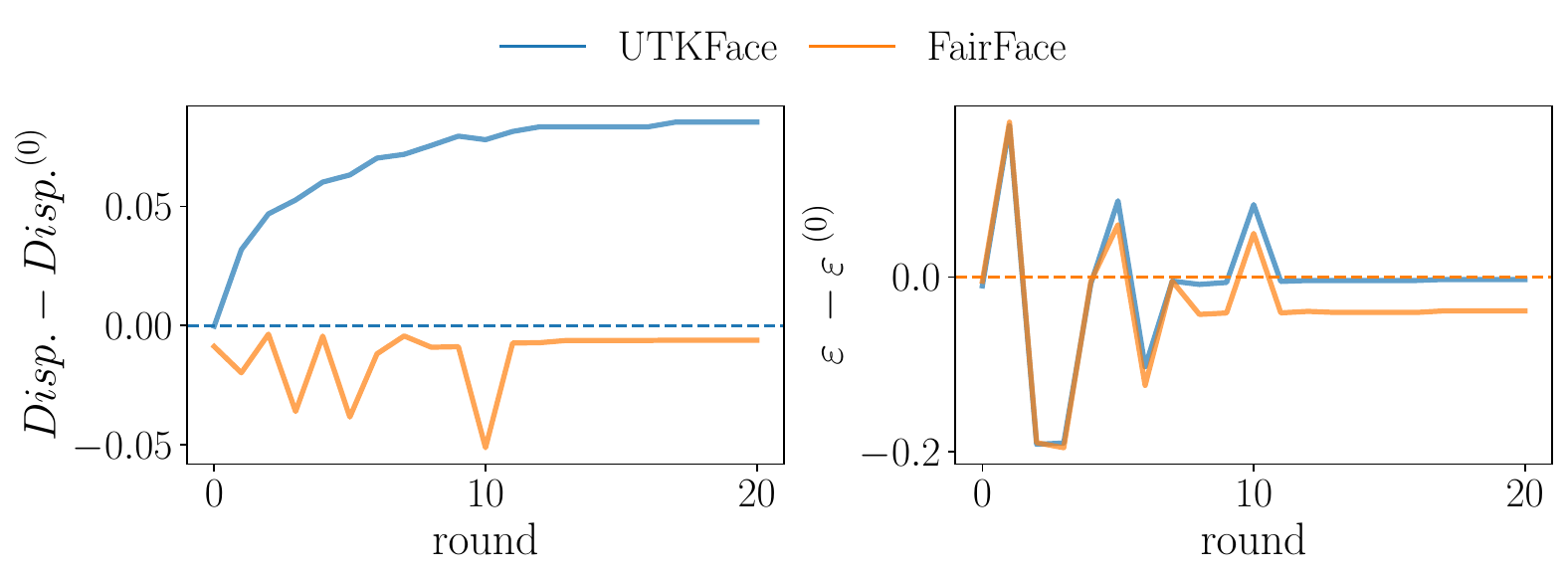}
    \caption{\textbf{Fairness regulator fails to enforce fairness constraint.} We simulate a game where agents have access to different datasets. Regulators observe that the fairness constraint is enforced on their dataset so do not assign any penalty. However, the constraint is not enforced on builder's dataset and they continue to relax the fairness parameter due to lack of penalty. }  
    \label{fig:RQ4_fairness_gap}
\end{figure}

In Section \ref{sec:results} Figure \ref{fig:two_datasets} we showed that agents can still run \pp when they have access to different datasets and regulators are able to enforce their constraints in most cases. However, there are scenarios where the convergence point fails to satisfy the trust constraints. We show an example of such case here in Figure \ref{fig:RQ4_fairness_gap}. This example follows the same setup as in Figure \ref{fig:two_datasets}, where regulators have access to FairFace and model builder has access to UTKFace. The convergence point of this game does not satisfy the fairness constraint on UTKFace but does on FairFace. This is because at the current privacy budget, higher fairness disparity gap is not achievable on FairFace. Further relaxing the fairness constraint input parameter does not lead to larger fairness gap anymore. Since regulators only have access to FairFace, they observe that the fairness gap is always below the threshold and thus do not assign any penalties. The model builder then continues to relax the fairness constraint input parameter without any consequences. In general, if two datasets have different ranges of fairness disparity gaps at each respective privacy budget and fairness regulators sets the fairness constraint close to the upper limit of fairness disparity gap of their dataset, the convergence point of the game may not satisfy their fairness constraint. 

\section{Proofs}
\label{sec:proofs}

\subsection{Convergence of \pp}
\label{thm:convergence}
We provide a proof sketch for the converge of \pp. We consider Algorithm 1 and its iterations up to convergence to an equilibrium $\boldsymbol{s}^*$; that is, we assume penalty scalers are not-adjusted mid-play for incentive design purposes (See discussion in Section 5). This does not reduce the generality of our claims here, since every adjustment of penalty scalers would force a new run of the algorithm which as we will show is convergent.

Furthermore, we instantiate the SpecGame which ParetoPlay simulates using the FairPATE learning framework of [44]. Apart from the particularities of each learning framework reflected in their loss terms, the analysis presented here should apply to other learning frameworks (such as DP-SGD methods).

We assume that the market is regulator-led, meaning that the regulators have already chosen guarantee levels  $\varepsilon_0$ and $\gamma_0$ . Since $C_*$ and $\lambda_*$ are both scalers, W.L.O.G., we assume $\lambda_\text{fair} = \lambda_\text{priv} = 1$ , then

\begin{align}
  F := L_\text{build}(\gamma, \varepsilon)=\ell_{\text {build }}(\gamma, \varepsilon) +C_{\text {fair}} \ell_{\text {fair }}(\gamma) +C_{\text {priv}} \ell_{\text {priv }}(\varepsilon)
\end{align}

For a builder using PATE-based methods, $\ell_\text{build}$ is the PATE-student average loss which is a typical neural network training loss (such as cross-entropy). 
  
The Pareto frontier is a trade-off function between acc-priv-fairness; so it is enough to be looking at $\ell_\text{build}(\gamma, \varepsilon)$ exclusively. We can show that Algorithm 1 is similar to minimization with subgradient methods with square summable but not summable step sequences~\citep{boyd_subgradient_2003}. If we can claim (*) is $L$-smooth, then we can use the standard argument in~\citet[Section 2]{boyd_subgradient_2003} to show convergence.

{
\renewcommand{\thetheorem}{\ref{thm:cne}}
\begin{theorem}
\pp recovers a Correlated Nash Equilibrium of the \specg. 
\end{theorem}
\addtocounter{theorem}{-1}
}

\begin{proof}
Assuming all agents play on the Pareto frontier (\Cref{alg:paretoplay}), we need to show that there is no incentive to deviate from playing on the Pareto frontier.

Assume that Builder (B) reports a $\omega_r$ that is not on the PF.  Assume there that there exists some $\omega^*$ on the PF, this means that $\omega^*$ Pareto dominates $\omega_r$ : it is at least as good in all objectives, and better at least in one.  We first note that reporting $\omega_r$ where $\ell_{acc}(\omega_r) > \ell_{acc}(\omega^*)$ is irrational (in the game theoretic sense that it increases the agent's cost instead of reducing it) and thus never a best response for B. So we can only cases where it holds that either $\ell_{acc}(\omega_r) < \ell_{acc}(\omega^*)$ and $\ell_{fair}(\omega_r) > \ell_{fair}(\omega^*)$ , or $\ell_{acc}(\omega_r) < \ell_{acc}(\omega^*)$ and $\ell_{priv}(\omega_r) > \ell_{priv}(\omega^*)$ , or both hold. But every agent in Pareto Play, re-calculates its Pareto frontier as a first step (Line 2 in Alg. 1). Assume, if at time $t-1$ ,  $B$ adds  $\omega_r$ to $R^{(t)}$ . At time $t$ , the regulator would re-calculate its PF; but since $\omega_r$ is not on the PF, either a) some other $\omega_*$ already exists in $R^{(t)}$ which dominates  $\omega_r$ , and therefore $\omega_r$ never appears in the rest of the regulators round; or b) if no such $\omega_*$ exists, the regulator will use  $\omega_r$ as initialization, do not change the penalty scale in Line 6 (again because it would be irrational for B to report a $\omega_r$ which would cause a penalty), and take a step on the Pareto frontier to improve the corresponding regulator loss. At this point, depending on which objective value was under-reported by B, the regulator would either be able to find an $\omega_*$ that Pareto dominates $\omega_r$ —at which point  $\omega_r$ again is effectively removed from the PF calculations—or the next regulator is going to make a gradient step and find the appropriate $\omega_*$ that Pareto dominates the misreported $\omega_r$ . In the worst-case where we lose gradient information (in a boundary condition, or near an inflection point), we note that every agent trains a model in the Calibration phase (line 10). At this point, with a near 0 gradient step, $\omega_* \approx \omega_r$ is reevaluated by one of the regulators, which ensures that $\ell_{priv}$ and/or $\ell_{fair}$ values are corrected, which again leads to exclusion of $\omega_r$ from the Pareto frontier. Therefore, we have shown that there is no incentive to play a Pareto inefficient solution.
\end{proof}
\section{Pareto Frontiers}
\label{app:pareto-frontiers}

\begin{figure}
\centering
    \begin{subfigure}{0.5 \linewidth}
    \includegraphics[width=0.9\linewidth]{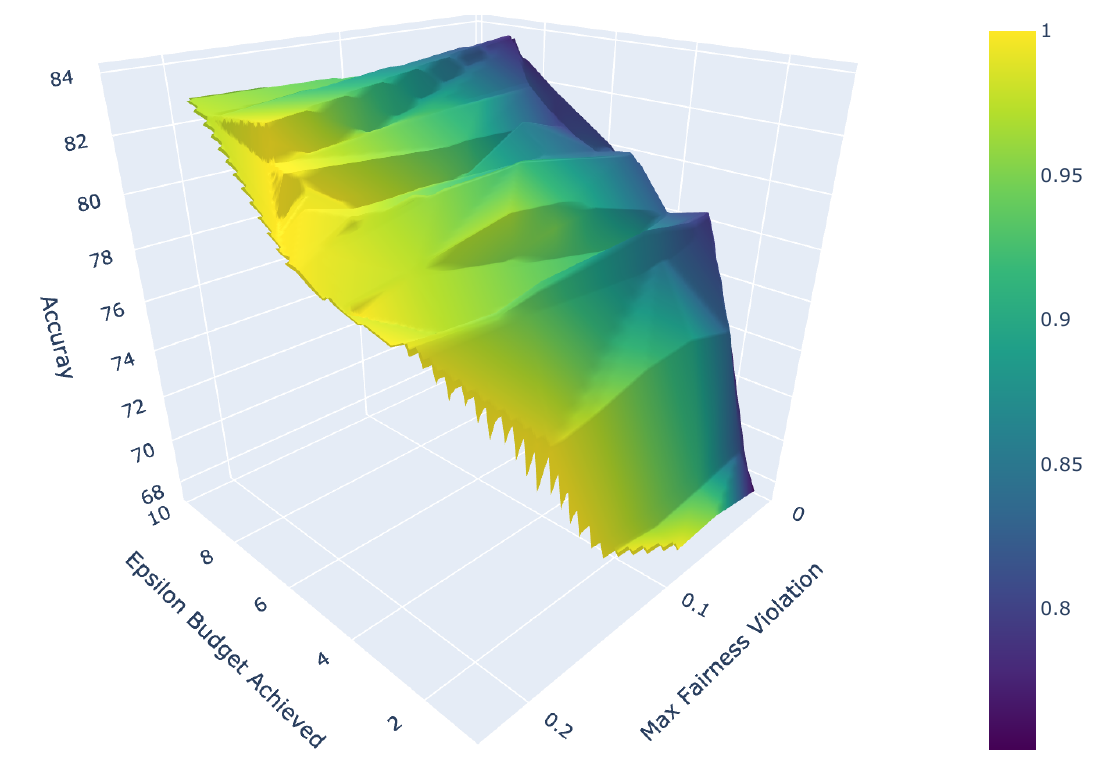}
    \caption{FairPATE Pareto Surface on UTKFace}
    \end{subfigure}
    \begin{subfigure}{0.4 \linewidth}
    \includegraphics[width=0.9\linewidth]{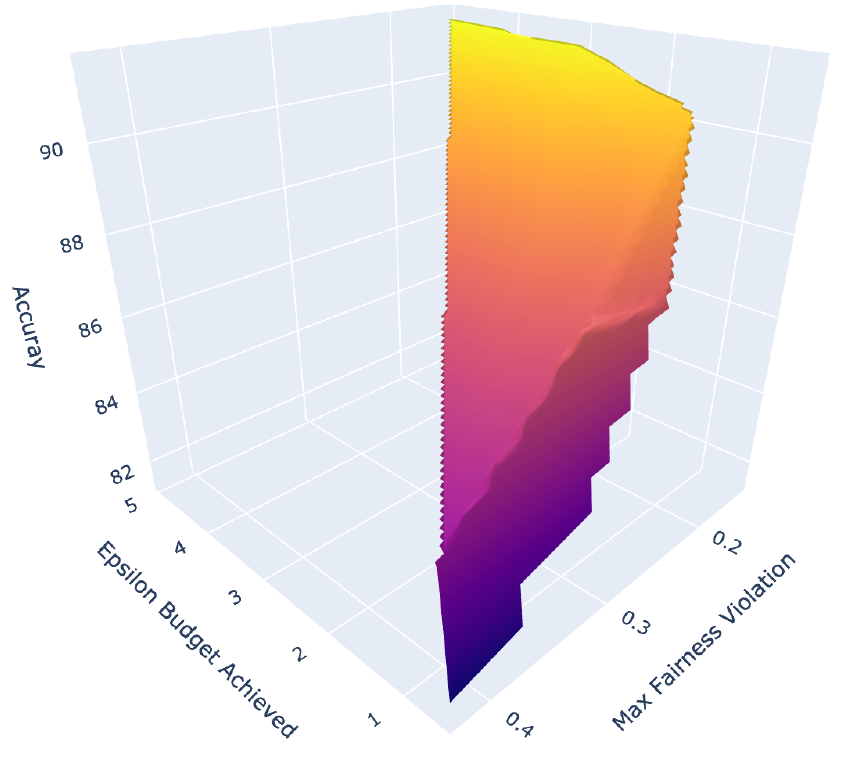}
    \caption{DPSGD-G.-A. Pareto Surface on MNIST}
    \end{subfigure}
    \caption{Pre-computed Pareto Frontier surfaces.}
    \label{fig:pf_surfaces}
\end{figure}

In~\Cref{fig:pf_surfaces}, we highlight the Pareto frontiers over which \pp is played that are experimental results in~\Cref{sec:results} demonstrate.

\end{document}